\documentclass[journal]{IEEEtran}

\ifCLASSINFOpdf
\else
\fi

\usepackage{amsmath}
\usepackage{amsfonts}
\usepackage{array}
\usepackage[pagebackref,breaklinks,colorlinks]{hyperref}
\usepackage{graphicx}
\usepackage{multirow}
\begin{document}

\title{TTVFI: Learning Trajectory-Aware Transformer for Video Frame Interpolation}

\author{Chengxu~Liu,
        Huan~Yang,
        Jianlong~Fu,
        Xueming~Qian
    \thanks{This work was done while Chengxu Liu was a research intern at Microsoft Research Asia.}
	\thanks{Chengxu Liu is with the School of Information and Communication Engineering, Xi’an Jiaotong University, Xi’an 710049, China (e-mail: liuchx97@gmail.com).}
	\thanks{Huan Ynag and Jianlong Fu are with  Microsoft Research (e-mail: huayan@microsoft.com; jianf@microsoft.com).}
	\thanks{Xueming Qian is with the Ministry of Education Key Laboratory for Intelligent Networks and Network Security, School of Information and Communication Engineering, and  SMILES LAB, Xi’an Jiaotong University, Xi’an 710049, China. (*Corresponding author, qianxm@mail.xjtu.edu.cn).}
	}


\maketitle

\begin{abstract}
Video frame interpolation (VFI) aims to synthesize an intermediate frame between two consecutive frames. 
State-of-the-art approaches usually adopt a two-step solution, which includes 1) generating locally-warped pixels by flow-based motion estimations, 2) blending the warped pixels to form a full frame through deep neural synthesis networks. 
However, due to the inconsistent warping from the two consecutive frames, the warped features for new frames are usually not aligned, which leads to distorted and blurred frames, especially when large and complex motions occur. 
To solve this issue, in this paper we propose a novel \textbf{T}rajectory-aware \textbf{T}ransformer for \textbf{V}ideo \textbf{F}rame \textbf{I}nterpolation (TTVFI). In particular, we formulate the warped features with inconsistent motions as query tokens, and formulate relevant regions in a motion trajectory from two original consecutive frames into keys and values. Self-attention is learned on relevant tokens along the trajectory to blend the pristine features into intermediate frames through end-to-end training. Experimental results demonstrate that our method outperforms other state-of-the-art methods in four widely-used VFI benchmarks. Both code and pre-trained models will be released at \href{https://github.com/researchmm/TTVFI.git}{https://github.com/researchmm/TTVFI}.

\end{abstract}


\IEEEpeerreviewmaketitle

\section{Introduction}

\begin{figure*}[!t]
\centering
\includegraphics[width=0.98\linewidth]{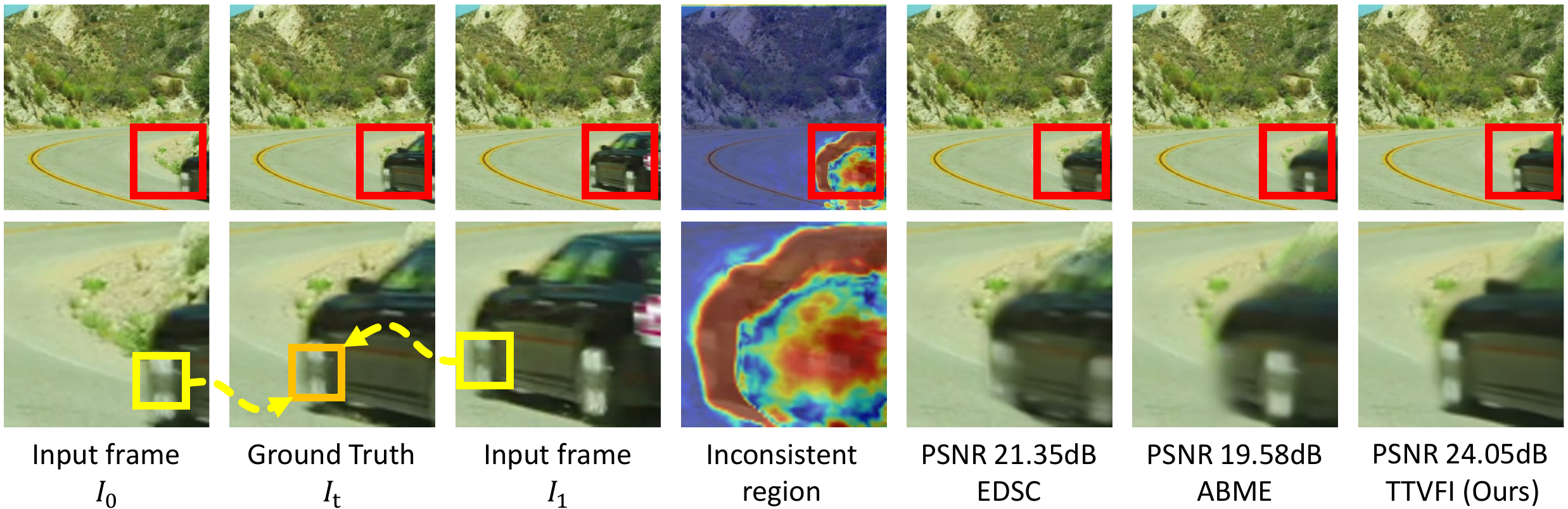}
\caption{A comparison between TTVFI and other SOTA methods: EDSC~\cite{cheng2021multiple} and ABME~\cite{park2021asymmetric}. Zoom in to see better visualization (indicated by red). TTVFI focuses on the inconsistent motion regions (indicated by warmer color), attention is learned on relevant tokens along the trajectory (indicated by yellow) to blend the pristine features into intermediate frames (indicated by orange).}
\label{fig:Teaser}
\end{figure*}

\IEEEPARstart{V}{ideo} frame interpolation (VFI) aims to synthesize non-existent frames between two consecutive frames. It is a fundamental problem in computer vision and can be applied to numerous applications, including slow-motion video generation~\cite{jiang2018super}, frame rate upconversion~\cite{bao2018high}, video compression~\cite{wu2018video}, and view synthesis~\cite{flynn2016deepstereo}. From a methodology perspective, unlike other image/video restoration tasks that usually recover enhanced images/videos from low-quality visual information on spatial dimensions, VFI tasks pay more attention to exploiting temporal motion information and synthesizing high-quality texture details in interpolated frames. As shown in Fig.~\ref{fig:Teaser}, if detailed textures to recover the target frame can be discovered and leveraged at adjacent frames, video qualities can be greatly enhanced.

Recently, classical frame interpolation algorithms synthesize the interpolated results either by predicting the blending kernels~\cite{cheng2021multiple,lee2020adacof,niklaus2017video2,niklaus2017video} or with help of motion estimation networks~\cite{bao2019depth,niklaus2018context,niklaus2020softmax,park2020bmbc,park2021asymmetric}. The former makes attempts to predict the blending kernels, and the interpolated result is obtained by filtering operation. However, the kernel size directly restricts the motion that the model can capture. Capturing larger motions with larger kernel size (e.g., $51\times 51$ in \cite{niklaus2017video}) results in heavy memory and computation cost. For the latter, benefiting from significant progress of motion estimation~\cite{meister2018unflow,sun2018pwc}, the typical frame interpolation algorithms use auxiliary of optical flow to synthesize the interpolated results, such as DAIN~\cite{bao2019depth}, BMBC~\cite{park2020bmbc}, and ABME~\cite{park2021asymmetric}. Nevertheless, the accuracy of the motion field and the manner of intermediate frame synthesis remain the great challenges that limit the effectiveness of VFI.

In particular, to solve this challenge, recent years have witnessed an increasing number of advanced algorithms~\cite{bao2019depth,niklaus2018context,niklaus2020softmax,park2020bmbc,park2021asymmetric} to investigate the effects of motion field (i.e.,~optical flow) in video frame interpolation. Typical algorithms~\cite{park2020bmbc,park2021asymmetric,xu2019quadratic} assume some pre-defined motion patterns (e.g., uniform, asymmetric motion) to estimate optical flow and input the warped frames obtained via bi-directional flow-based warping to synthesis network. 
However, there are still some problems as follows:
1)~The synthesis network focuses on achieving overall interpolation averaged over all regions of intermediate results. 
For some challenging scenes (e.g.,~fast-moving, turn around), the pre-defined motion patterns may produce inaccuracy or inconsistent motion fields (e.g.,~the symmetric bilateral motion and asymmetric bilateral motion in Fig.~\ref{fig:motion}). 
2)~The flow-based warping will produce inaccurate texture synthesis in inconsistent motion region (e.g.,~the inconsistent region in Fig.~\ref{fig:Teaser}), which is common and necessary in VFI.
Therefore, such designs lack a necessary design to improve the interpolation results in important regions and may produce distortion and blurring (e.g.,~the ABME~\cite{park2021asymmetric} in Fig.~\ref{fig:Teaser}). A more promising solution is to explore a proper synthesis network for generating intermediate results by introducing pristine features of the original input frames.

Besides, inspired by the recent significant progress of Transformer~\cite{vaswani2017attention} in video restoration~\cite{zeng2020learning,cao2021video,shi2021video,liu2022learning}, VSR-Transformer~\cite{cao2021video} and TTVSR~\cite{liu2022learning} propose to use Transformer to generate the enhanced and high-resolution object in recovered video. In VFI, VFIT~\cite{shi2021video} proposes to use Transformer to extract deep hierarchical features, and predict the blending kernels for interpolating results. However, this method benefits from the long-range dependent learning capability of the Transformer itself and has not exploited the potential of the attention mechanism in object modeling and improving interpolation results. Therefore, in VFI, utilizing Transformer to synthesize high-quality texture details and pleasing interpolation results remains a great challenge.

In this paper, we propose a novel Trajectory-aware Transformer to achieve more accurate and effective feature learning in Video Frame Interpolation (TTVFI), as shown in Fig.~\ref{fig:transformer}. The key insight is to focus on the regions with inconsistent motion (e.g., the inconsistent region in Fig.~\ref{fig:Teaser}), and allow features to be learned from the original input frames through the attention mechanism. 
In particular, we propose a consistent motion learning component in trajectory-aware Transformer at first, as shown in Fig.~\ref{fig:cml}, to obtain motion fields, which can be used to generate a group of inter-frame motion trajectories. 
Then, the trajectories and motion fields are used to formulate the two kinds of visual tokens. 
They come from the original input frames and warped frames and learn on the relevance of them in regions with inconsistent and consistent motion, respectively.
Finally, once the tokens have been obtained, TTVFI learns relevant features by calculating self-attention in regions with inconsistent and consistent motion.
The output of TTVFI can be stacked in multi-scale to further boost feature representation of intermediate results. 

Compare with VFIT~\cite{shi2021video} that use Transformer to predict the blending kernels for interpolating results. TTVFI selects features from the input frames along the trajectory and synthesize richer textures in a trajectory-based way. This manner exploits the potential of feature restoration in the synthesis network and improves interpolation results through well-designed visual tokens along the motion trajectory.

Our contributions are summarized as follows:
\begin{itemize}
    \item We propose a novel trajectory-aware Transformer, which enables more accurate features learning of synthesis network by introducing Transformer into VFI tasks. Our method focuses on regions of video frames with motion consistency differences and performs attention with two kinds of well-designed visual tokens along the motion trajectory.
    \item We propose a consistent motion learning module to generate the consistent motion in trajectory-aware Transformer, which is used to generate the trajectories and guide the learning of the attention mechanism in different regions.
    \item Extensive experiments demonstrate that the proposed TTVFI can outperform existing state-of-the-art methods in four widely-used VFI benchmarks.
\end{itemize}

The rest of the paper is organized as follows. Related work is reviewed in Sec.~\ref{rw}. The proposed trajectory-aware Transformer is elaborated in Sec.~\ref{tt}. Experimental evaluation, analysis, and ablation study are presented in Sec.~\ref{e}. The discussion of the related parameters and component are presented in Sec.~\ref{d}. The limitations and failure cases are elaborated in Sec.~\ref{l}. Finally, we conclude this work in Sec.~\ref{c}.

\section{Related Work}
\label{rw}
In this section, we mainly introduce the related work on video frame interpolation. Additionally then, we give a brief overview of visual Transformer and their application in various fields. 

\subsection{Video Frame Interpolation}
Video frame interpolation is a classical problem in various image processing and computer vision tasks~\cite{flynn2016deepstereo,jiang2018super,wu2018video,shen2020video,zhao2019enhanced}. In this section, we focus on recent VFI algorithms, which can be classified into two paradigms: kernel-based~\cite{cheng2021multiple,lee2020adacof,niklaus2017video2,niklaus2017video} methods and flow-based~\cite{bao2019depth,bao2019memc,gui2020featureflow,liu2019deep,liu2017video2,niklaus2018context,niklaus2020softmax,park2020bmbc,park2021asymmetric,xue2019video} methods. 

\subsubsection{Kernel-based video interpolation}
The kernel-based methods make attempts to estimate the blending kernels using CNNs~\cite{niklaus2017video,niklaus2017video2} or deformable convolutions~\cite{cheng2020video,cheng2021multiple,dai2017deformable}, and the interpolated result is obtained by filtering operation. Typically, AdaConv~\cite{niklaus2017video} and SepConv~\cite{niklaus2017video2} predict spatially-adaptive and separable interpolation kernels respectively to aggregate each pixel from the neighborhood. DSepConv~\cite{cheng2020video} and EDSC~\cite{cheng2021multiple} propose adaptively estimate kernels using deformable separable convolution to extend the receptive field of the pre-defined kernel and focusing on more relevant pixels. To solve the degrees of freedom limitations in complex motions, AdaCoF~\cite{lee2020adacof} propose to estimates both kernel weights and offset vectors for each pixel. 

However, the kernel size directly restricts the motion that the model can capture. Capturing larger motions with larger kernel size (e.g., $51\times 51$ in \cite{niklaus2017video}) results in heavy memory and computation cost. 

\begin{figure*}[t!]
\centering
\includegraphics[width=1.0\linewidth]{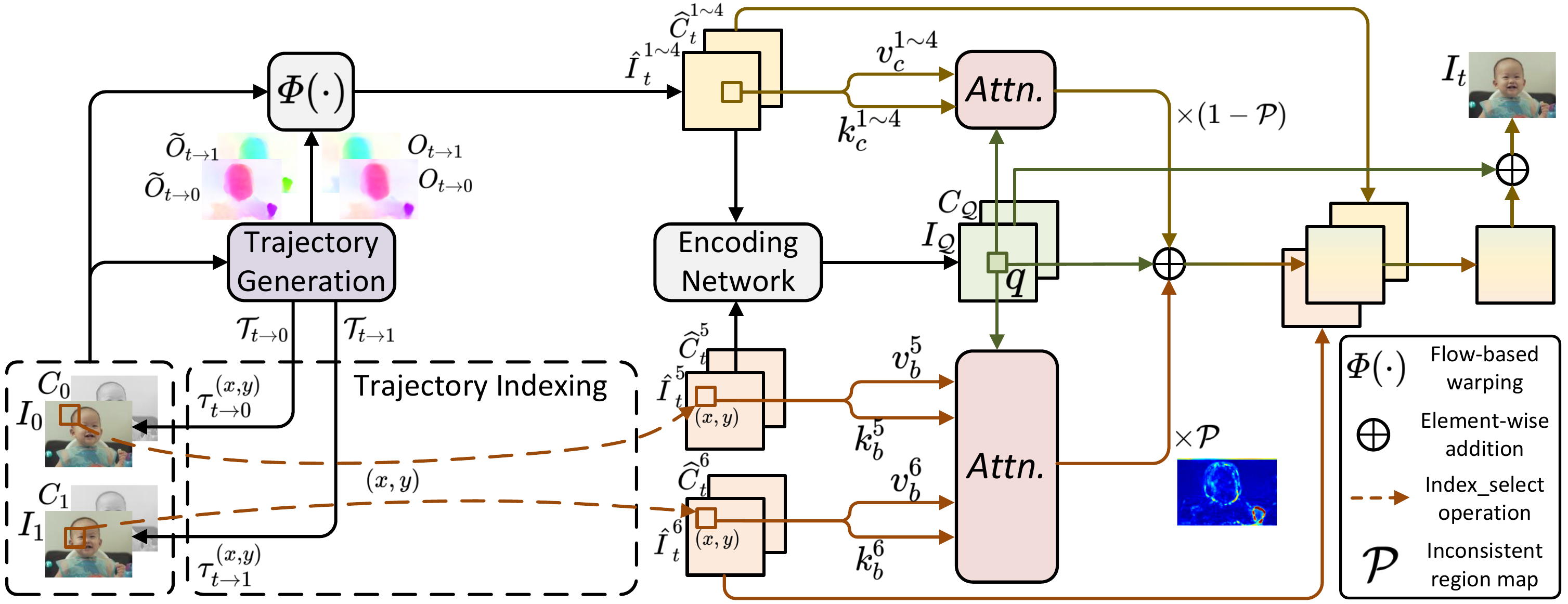}
\caption{The overview of TTVFI. $I_{0},I_{1}$ and $C_{0},C_{1}$ are the input frames and contextual features, respectively. $\widetilde{O}_{t\to0}, \widetilde{O}_{t\to1}$ and $O_{t\to0}, O_{t\to1}$ indicate two kinds of motion fields. $\tau^{(x,y)}_{t\to 0}$ and $\tau^{(x,y)}_{t\to 1}$ are elements of the trajectories set $\mathcal{T}_{t\to 0}$ and $\mathcal{T}_{t\to 1}$ at the start point $(x,y)$, respectively. $q$, $k$, and $v$ indicate the query, key, and value, respectively. $\mathcal{P}$ indicates the inconsistent region map. $I_{t}$ indicates the output intermediate frame.}
\label{fig:transformer}
\end{figure*}

\subsubsection{Flow-based video interpolation}
Unlike relying on kernel estimation, the flow-based methods have been developed most actively and usually consist of two steps: 1) warping the input frames based on the optical flow from the motion estimation network, 2) blending the warped frames through the synthesis network. The flow-based methods focus on generating more accurate motion to warp the input frames, and contain two algorithms using forward warping~\cite{niklaus2020softmax} and backward warping~\cite{bao2019depth,gui2020featureflow,niklaus2018context,park2020bmbc,park2021asymmetric}. Typically, SoftSplat~\cite{niklaus2020softmax} proposes softmax splatting to address the conflict of mapping multiple pixels to the same target location in forward warping, but suffers from holes pixels. For the methods using backward warping, CtxSyn~\cite{niklaus2018context} presents a context-aware synthesis approach to effectively blend the two warped frames. DAIN~\cite{bao2019depth} introduces the depth information to deal with the holes or overlay caused by occlusion. FeatureFlow~\cite{gui2020featureflow} proposes to predict the optical flow of features to handle the interpolation of complex dynamic scenes. Further, to estimate the motion more accurately, BMBC~\cite{park2020bmbc} and ABME~\cite{park2021asymmetric} pre-define symmetric and asymmetric bilateral motion patterns to estimate the optical flow between video frames. All these methods reconstruct intermediate frame by blending the warped frames through the synthesis network.

However, for some challenging scenes, the pre-defined motion patterns may produce inaccurate or inconsistent motion fields, resulting in distortion and blurring.
Besides, these methods focus on achieving overall interpolation averaged over all regions of the intermediate result by the synthesis network and lack a necessary design for improving the interpolation result in important regions.

\subsection{Visual Transformer}

Recently, due to its advanced learning capabilities, Transfomer~\cite{vaswani2017attention} as a new attention-based paradigm for modeling relationships between visual tokens in many computer vision tasks, such as image classification~\cite{dosovitskiy2020image,liu2021swin}, object detection~\cite{carion2020end} and so on. Typically, in video super-resolution, VSR-Transofmer~\cite{cao2021video} learns the fine texture from video frames through Transformer. STTN\cite{zeng2020learning} uses Transformer to video inpainting by searching missing contents from reference frames. In VFI tasks, benefiting from the long-range dependence learning capability of the Transformer, VFIT~\cite{shi2021video} predicts the blending kernels for achieving interpolation. In general, Transformer can be well-used for visual object recovery in the tasks of video reconstruction. 

Besides, Motionformer~\cite{patrick2021keeping} proposes trajectory attention that aggregates information along implicitly determined trajectory to video action recognition. TTVSR~\cite{liu2022learning} also proposes trajectory-aware Transformer to enable effective long-range spatio-temporal learning in videos super-resolution tasks. There are different from the field and the implication of trajectory in our work. In this paper, we propose a novel trajectory-aware Transformer that improves interpolated results by performing attention in different regions of the frame with different pre-defined visual tokens along the motion trajectories.

\section{Trajectory-aware Transformer}
\label{tt}
\subsection{Overview}

Existing works~\cite{bao2019depth,niklaus2018context,park2021asymmetric} warp the input frames by the optical flow with pre-defined motion patterns, and lack a necessary design for the important synthesis network. 
Therefore, we propose the trajectory-aware Transformer to mitigate the distortion and blur caused by inconsistent warping and synthesize the interpolation results.

As shown in Fig.~\ref{fig:transformer}, TTVFI takes two successive frames $I_{0}$, $I_{1}$ and extracted context feature $C_{0}$, $C_{1}$ as input, and generates an intermediate frame $I_{t}, t\in (0,1)$. Specifically, we first propose a trajectory generation module to obtain the motion field $\mathcal{Q}$ and trajectory $\mathcal{T}$ between two successive video frames. Then, we use the flow-based warping $\Phi(\cdot)$ and trajectory indexing $Idx(\cdot)$ to generate two different features from different sources separately, and formulate them into two kinds of visual tokens by encoding network, named as consistent tokens $v_{c}$, $k_{c}$ and boundary tokens $v_{b}$, $k_{b}$. Finally, we perform trajectory-aware attention ${A}_{traj}(\cdot)$ in regions with different motion consistency (indicated by $\mathcal{P}$). The attention results are fed into a feed-forward network ${FFN}(\cdot)$ consisting of residual blocks (omitted for brevity in Fig.~\ref{fig:transformer}), and output the feature of the intermediate frame $I_{t}$. In summary, the trajectory-aware Transformer ${T}_{traj}(\cdot)$ can be formulated as:
\begin{equation}
\begin{aligned}
&\text{T}_{traj}(\mathcal{Q}, \mathcal{K}, \mathcal{V})\\
&=\text{FFN}(\text{A}_{traj}((\mathcal{Q}, \mathcal{K}_{c}, \mathcal{V}_{c}), (\mathcal{Q}, \mathcal{K}_{b}, \mathcal{V}_{b}), \mathcal{T})+\mathcal{Q}) , 
\end{aligned}
\label{equ:OTVFI}
\end{equation}
where $(\mathcal{K}_{c},\mathcal{V}_{c})$ and $(\mathcal{K}_{b},\mathcal{V}_{b})$ indicate the consistent tokens set and boundary tokens set, respectively. $\mathcal{T}$ is the motion trajectory. $\mathcal{Q}$, $\mathcal{K}$, $\mathcal{V}$ indicate the generic element queries, keys and values entered into Transformer. 
Note that we stack trajectory-aware Transformer on multiple scales to facilitate the learning of features. Here, we describe this structure only at one scale for brevity.

\subsection{Trajectory Generation}
\label{tg}
To alleviate the effects of inconsistent warping. We first estimate the consistent motion with a proposed consistent motion learning component. Then the consistent motion is further used to generate the motion trajectories of tokens.

\subsubsection{Consistent motion learning component}
\label{cmlc}
In video frame interpolation, since the intermediate frame $I_{t}$ is not available, it is not possible to directly obtain the motion field between the input frames $I_{0}$, $I_{1}$ and the intermediate frame $I_{t}$. 

Existing methods~\cite{park2020bmbc,park2021asymmetric} estimate an approximated motion by pre-defining some specific motion patterns (i.e., symmetric bilateral motion and asymmetric bilateral motion), which can be represented as:
\begin{equation}
\begin{split}
&O_{t\to0} = -t(1-t)\cdot O_{0\to1} + t^2\cdot O_{1\to0} ,\\
&O_{t\to1} = (1-t)^2\cdot O_{0\to1} - t(1-t)\cdot O_{1\to0} ,
\label{equ:1}
\end{split}
\end{equation}
where $O_{t\to0}$ and $O_{t\to1}$ are approximated by combining $O_{0\to1}$ and $O_{1\to0}$, which indicate the motion field between $I_{0}$ and $I_{1}$. For fair comparison, we follow previous works~\cite{bao2019depth,park2020bmbc,park2021asymmetric} to obtain $O_{0\to1}$ and $O_{1\to0}$ by PWC-Net~\cite{sun2018pwc}. However, the approximated motion ignores the consistency between two consecutive frames and leads to incorrect results for challenging scenes. 
\begin{figure}[!t]
\centering
\includegraphics[width=1.0\linewidth]{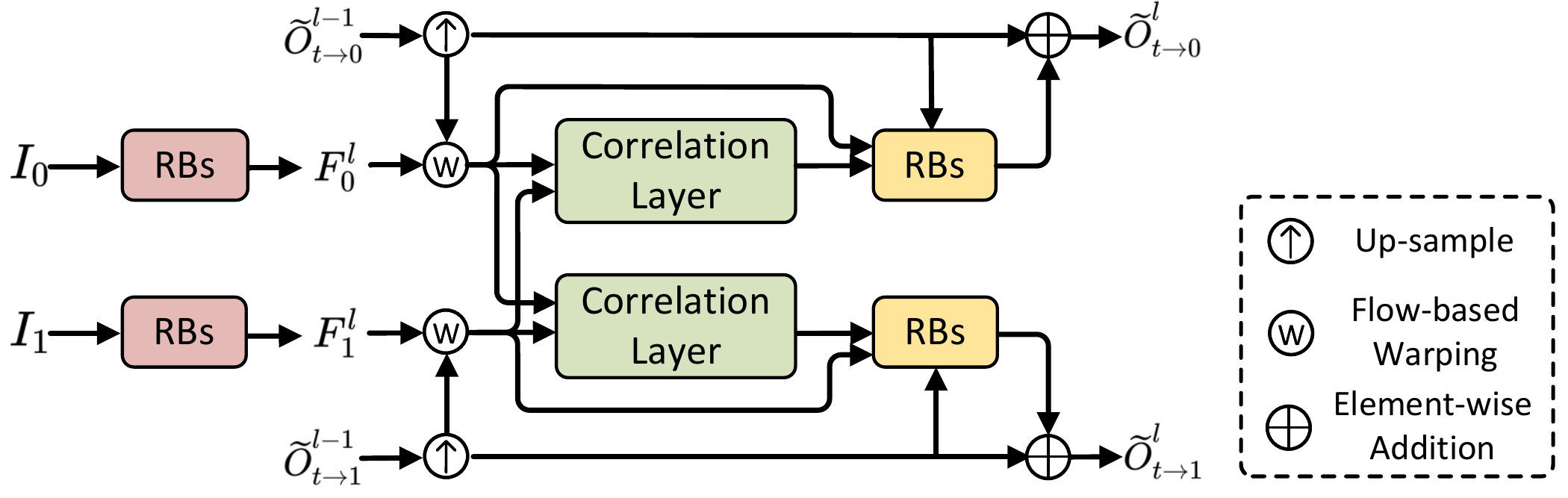}
\caption{The architecture of the consistency motion learning component.}
\label{fig:cml}
\end{figure}

Therefore, as shown in Fig.~\ref{fig:cml}, we propose a consistent motion learning component based on the approximated motion, which is integrated into the two largest scales of the PWC-Net~\cite{sun2018pwc}. The initial input of the component comes from the Equ.~\ref{equ:1}. The output of the component in the last scale is the consistent motion $\widetilde{O}_{t\to0}$ and $\widetilde{O}_{t\to1}$ with opposite directions simultaneously. Specifically, $\widetilde{O}_{t\to0}^{l-1}$ and $\widetilde{O}_{t\to1}^{l-1}$ indicate the motion from level $l-1$, it is up-sampled to warp the features $F^{l}_{0}$ and $F^{l}_{1}$ from level $l$. The matching costs of the two warped features are then computed in the correlation layer~\cite{sun2018pwc} (indicated by green) in an interactive way. Then, for getting $\widetilde{O}_{t\to0}^{l}$, we use the output cost volume from correlation layer, the warped feature from $F_{0}^{l}$ and the up-sampled motion from $\widetilde{O}_{t\to0}^{l-1}$ as input to generate the residual field. Finally, the residual field is added to the up-sampled $\widetilde{O}_{t\to0}^{l-1}$ to yield the $\widetilde{O}_{t\to0}^{l}$. $\widetilde{O}_{t\to1}^{l}$ can be obtained in the same way. The stacked residual block (indicated by red and yellow) is the same as the residual block used in PWC-Net. 

The core advantage of this component is that the two input optical flow in opposite directions can be optimized with each other and output simultaneously. Compared with approximated motion, the consistent motion has better temporal coherence, which helps in better trajectory generation in the following part. 

\subsubsection{Trajectory formulation}
The trajectories $\mathcal{T}_{s\to e}$ in our approach can be formulated as a set of trajectories, in which each trajectory $\tau^{(x,y)}_{s\to e}$ contains two coordinates. The start point is associated with the coordinate of the token at position $(x,y)$ at time $s$ and the endpoint is associated with the coordinate of the token at time $e$. They can be defined as:
\begin{equation}
\begin{aligned}
\mathcal{T}_{s\to e} =&\{\tau^{(x,y)}_{s\to e}| \:x \in \{1,\dots, H\}, y\in \{1,\dots, W\}\},\\
&\tau^{(x,y)}_{s\to e} =\langle (x, y),(x_{s\to e}, y_{s\to e}) \rangle,\\
\end{aligned}
\label{equ:deftraj}
\end{equation}
where $(x_{s\to e}, y_{s\to e})$ represents the coordinate transformation of the token at position $(x,y)$ from time $s$ to $e$. $H$ and $W$ represent the height and width of the features, respectively. Specifically, the trajectories $\mathcal{T}_{s\to e}$ can be calculated by:
\begin{equation}
\mathcal{T}_{s\to e} = \gamma(M + \widetilde{O}_{s\to e}),
\label{equ:motion2traj}
\end{equation}
where $M$ represents a two-dimensional meshgrid matrix~\footnote{Where the matrix index is equal to the element (i.e., $M(x,y)=(x,y)$).} of the same size as consistent motion $\widetilde{O}_{s\to e}$. $\gamma(\cdot)$ indicates the rounding operation to align the coordinates of the tokens at the endpoint of the trajectories.

\subsection{Token Generation}
To build visual tokens from different sources separately, we first generate two different features, named as warped features and extracted features. Then they are used to build query, key, and value tokens respectively.

Specifically, the warped features $\widehat{C}_{t}^{k}, k \in \{1,2,3,4\}$ are obtained by bi-directional flow-based warping. For regions with consistent motion, the warped features from two consecutive frames are well aligned, which can be obtained by:
\begin{equation}
\begin{split}
&\widehat{C}_{t}^{1} = \Phi_{b}(O_{t\to0},C_{0}), \quad \widehat{C}_{t}^{2} = \Phi_{b}(O_{t\to1},C_{1}),\\
&\widehat{C}_{t}^{3} = \Phi_{b}(\widetilde{O}_{t\to0},C_{0}), \quad \widehat{C}_{t}^{4} = \Phi_{b}(\widetilde{O}_{t\to1},C_{1}),
\end{split}
\end{equation}
where $\Phi_{b}(\cdot)$ is the backward warping. $C_{0}$ and $C_{1}$ are the contextual features obtained from the input frames by two convolutional layers. The extracted features $\widehat{C}_{t}^{k}, k \in \{5,6\}$ are obtained by extracting the features of input frames along the trajectories. For regions with inconsistent motion, the extracted features can introduce the pristine features from the original input, which can be obtained by:
\begin{equation}
\begin{split}
&\widehat{C}_{t}^{5} = \text{Idx}(\mathcal{T}_{t\to 0},C_{0}), \quad \widehat{C}_{t}^{6} = \text{Idx}(\mathcal{T}_{t\to 1},C_{1}),\\
\end{split}
\end{equation}
where $\text{Idx}(\cdot)$ denotes the operation of trajectory indexing (i.e., $index\_select$ \footnote{The $index\_select$ function implemented in PyTorch.}).
The wraped frames $\widehat{I}_{t}^{k}, k \in \{1,2,3,4\}$ and extracted frames $\widehat{I}_{t}^{k}, k \in \{5,6\}$ can be obtained in the same way.

\subsubsection{Query} We build queries by the output feature from a proposed encoding network. Inspired by previous work~\cite{niklaus2018context,park2021asymmetric}, the encoding network can be split into a GridNet~\cite{fourure2017gridnet} to generate filters and a dynamic local convolution~\cite{jia2016dynamic} to output feature of intermediate frame. 

First, we use the GridNet to generate local blending filters by inputting 
all the features $\widehat{C}_{t}^{k}, k \in \{1,2,3,4,5,6\}$ and frames $\widehat{I}_{t}^{k}, k \in \{1,2,3,4,5,6\}$ obtained above. Then, the generated filters by the GridNet can be denoted as $H_{(x,y)}(i,j,k)$, where $(i,j,k)\in\{-1,0,1\}\times\{-1,0,1\}\times\{1,2,3,4,5,6\}$ is the local coordinate around $(x,y)$ in the features. The range of $(i,j)$ is dependent on the kernel size generated by GridNet. The dynamic local convolution uses the generated filters to yield the feature of intermediate frame $C_{\mathcal{Q}}$ by:
\begin{equation}
C_{\mathcal{Q}} = \sum^{6}_{k=1}\sum^{1}_{i=-1}\sum^{1}_{j=-1}H_{(x,y)}(i,j,k)\cdot \widehat{C}_{t}^{k}(x+i,y+j),
\label{equ:Q}
\end{equation}
where the coefficients are normalized by $\sum_{k}\sum_{i}\sum_{j}H_{(x,y)}(i,j,k)=1$ to ensure the magnitude of the pixels after convolution. By introducing the information from neighboring pixels, the convolution can compensate for the inconsistent motion to an extent. The intermediate frame $I_{\mathcal{Q}}$ also can be obtained in the same way. Finally, this feature and frame are fed into an embedding layer of one convolutional layer to build the queries. This process can be represented as:
\begin{equation}
\mathcal{Q} = E(\text{Concat}(C_{\mathcal{Q}}, I_{\mathcal{Q}})),
\label{equ:Q2}
\end{equation}
where $\text{Concat}(\cdot)$ and $E(\cdot)$ denote the concatenate operation and the embedding layer, respectively.

\subsubsection{Key and value} We formulate input frames into two kinds of visual tokens, named as consistent tokens and boundary tokens. 

In particular, consistent tokens $(\mathcal{K}_{c},\mathcal{V}_{c})$ focus on the regions where the motion is consistent and well coherent. Thus, consistent tokens can be accurately generated by the warped features $\widehat{C}_{t}^{k}, k \in \{1,2,3,4\}$ and frames $\widehat{I}_{t}^{k}, k \in \{1,2,3,4\}$. This process can be represented as:
\begin{equation}
\mathcal{K}_{c} =\mathcal{V}_{c} =  E(\text{Concat}(\widehat{C}_{t}^{k},\widehat{I}_{t}^{k})), k \in \{1,2,3,4\}.\\
\end{equation}
Boundary tokens $(\mathcal{K}_{b},\mathcal{V}_{b})$ focus on the regions with inconsistent motion, which mainly appear at the boundaries of moving instances. The inaccurate warping caused by inconsistent motion can destroy the pristine features in the original input frames. Therefore, we use the extracted features $\widehat{C}_{t}^{k}, k \in \{5,6\}$ and frames $\widehat{I}_{t}^{k}, k \in \{5,6\}$ to construct the boundary tokens. This process can be represented as:
\begin{equation}
\mathcal{K}_{b} = \mathcal{V}_{b} = E(\text{Concat}(\widehat{C}_{t}^{k},\widehat{I}_{t}^{k})), k \in \{5,6\}.\\
\end{equation}

Based on the two kinds of well-designed tokens, the model can perform attention mechanisms in different regions according to the motion consistency.

\subsection{Trajectory-aware Attention}
\label{ja}
To mitigate the distortion and blur caused by inconsistent motion, we further introduce an inconsistent region map $\mathcal{P}$ to guide the attention calculation.

\subsubsection{Inconsistent region map} 
The inconsistent region map $\mathcal{P}$ is the same size as the input frame and indicates a confidence measure of motion consistency for different regions. It can be obtained as follow:
\begin{equation}
\mathcal{P} = 2 \cdot \text{Sigmoid} (|\widetilde{O}_{t\to0}+\widetilde{O}_{t\to1}|) - 1,
\end{equation}
where $\text{Sigmoid}(\cdot)$ is the sigmoid function for normalization. The sum of the optical flows $|\widetilde{O}_{t\to0}+\widetilde{O}_{t\to1}|$ in opposite directions reflects the coherence of motion. For consistent regions, it can be completely offset and the value of $\mathcal{P}$ converges to 0. Conversely, for inconsistent motion, the value of $\mathcal{P}$ converges to 1. The purpose of the proposed inconsistent region map is to distinguish the inconsistent regions and guide the calculation of attention.

\subsubsection{Attention calculation} 
\label{ac}
The input of attention mechanism consists of queries $\mathcal{Q}$, consistent tokens $(\mathcal{K}_{c},\mathcal{V}_{c})$, boundary tokens $(\mathcal{K}_{b},\mathcal{V}_{b})$, and inconsistent region map $\mathcal{P}$. In the calculation process, we first compute the dot products of the query with all keys, divide each by scaling factor and apply a softmax function to obtain the weights on the values. Then, the output of the attention mechanism can be obtained by weighting the sum of two kinds of tokens with the obtained weights and $\mathcal{P}$. We compute the features of outputs as follow:
\begin{equation}
\begin{aligned}
&\text{A}_{joint}((\mathcal{Q},\mathcal{K}_{c},\mathcal{V}_{c}),(\mathcal{Q},\mathcal{K}_{b},\mathcal{V}_{b}),\mathcal{P}) \\
&= (1-\mathcal{P})\cdot S(\frac{\mathcal{Q}\mathcal{K}_{c}^{T}}{\sqrt{d_{k_{c}}}})\mathcal{V}_{c}+\mathcal{P}\cdot S(\frac{\mathcal{Q} \mathcal{K}_{b}^{T}}{\sqrt{d_{k_{b}}}})\mathcal{V}_{b},
\end{aligned}
\end{equation}
where $S(\cdot)$ denotes the softmax function. $d_{k_{c}}$ and $d_{k_{b}}$ denote the dimension of two kinds of keys. Besides, the tokens only produce a local position offset after the motion, so it is unnecessary and unrewarding in performing attention globally. Inspired by Swin  Transformer~\cite{liu2021swin}, we perform the attention mechanism inside each shifted window to reduce the computational cost. In each attention layer, we add the feed-forward network that consists of a convolutional layer of size $3\times 3$ and a PReLU~\cite{he2015delving} activation following it. The feed-forward network is applied to each position and considers the neighboring pixels to restructure the output feature of the trajectory-aware Transformer.

\subsubsection{Multi-scale fusion} 
\label{msf}
\begin{figure}[!t]
\centering
\includegraphics[width=1.0\linewidth]{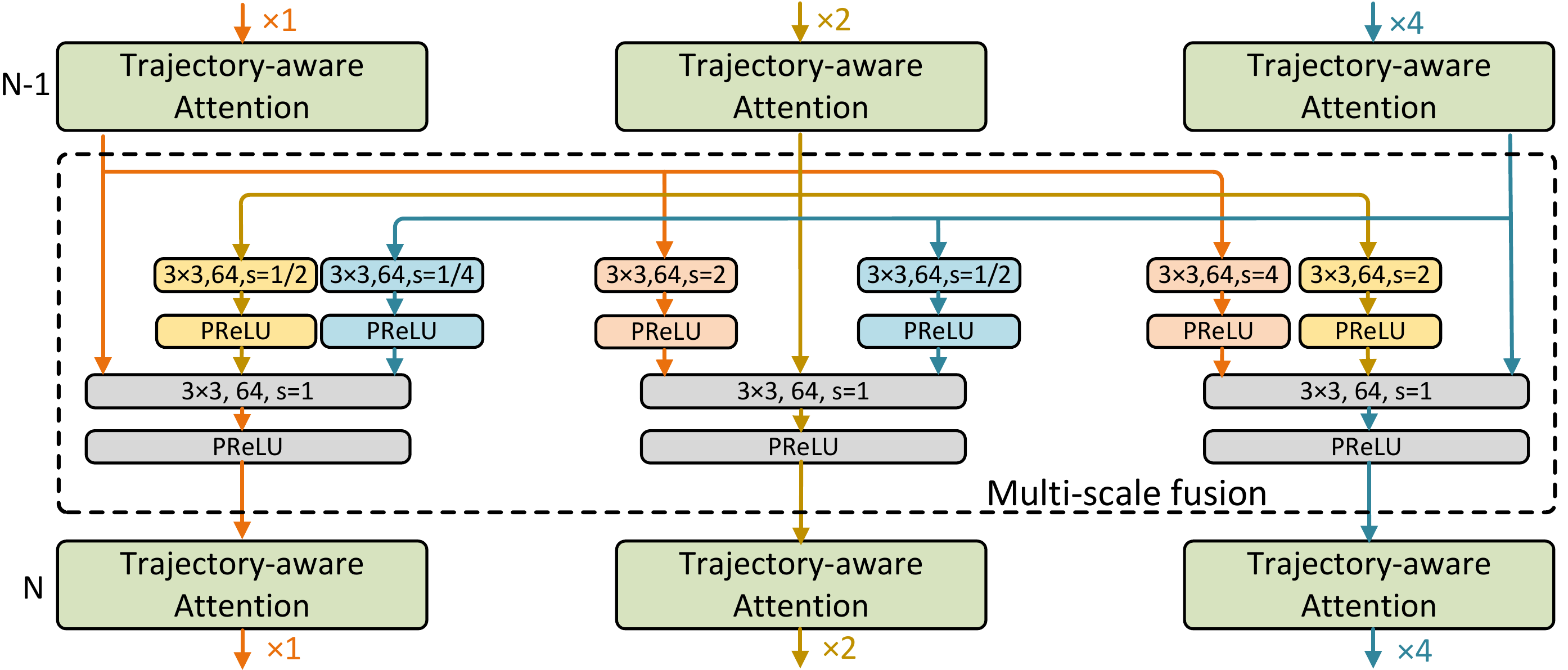}
\caption{The architecture of the multi-scale fusion.}
\label{fig:msf}
\end{figure}
In the previous works~\cite{yang2020learning,liu2021swin}, stacking transformers in multi-layer and multi-scale has been proven to be effective. Therefore, to boost the generated feature representation of intermediate results, we stack the proposed trajectory-aware Transformer in multi-scale (i.e. $\times1$, $\times2$, and $\times4$) to achieve a more powerful feature representation. Specifically, as shown in Fig.~\ref{fig:msf}, to facilitate the interaction of multi-scale features, we use a multi-scale fusion module (indicated by red) in the hierarchical structure. This design enables information at each scale to exchange with each other and fuse together in a cross-scale manner. The final fused feature generates the residual that is added to the intermediate frame $I_{\mathcal{Q}}$ obtained-above to output the final intermediate frame $I_{t}$.

In general, we exploit the potential of feature restoration in the synthesis network, which is neglected in video frame interpolation. By introducing the trajectory-aware Transformer, we perform the attention mechanism along the motion trajectory with well-designed visual tokens for inconsistent regions and enable the synthetic network to learn more accurate features.

\subsection{Training}
For fair comparisons, we follow existing works~\cite{gui2020featureflow,park2020bmbc,park2021asymmetric} to adopt a two-stage strategy to optimize our model. In stage one, we train the consistent motion learning component to obtain the motion fields. Then, we end-to-end train the whole model in stage two.

\subsubsection{Stage one}
To improve the consistency of motion between the consecutive frames and the robustness of optical flow to illumination changes, we define the consistent loss $L_{con}$ and the census loss $L_{cen}$ as follow:
\begin{equation}
\begin{aligned}
L_{con} &= \varphi(I_{t}^{GT}-\widehat{I}_{0\to t}) + \varphi(I_{t}^{GT}-\widehat{I}_{1\to t})\\
&+\varphi(I_{1}-\Phi_{b}(O_{1\to t},\widehat{I}_{0\to t}))\\
&+ \varphi(I_{0}-\Phi_{b}(O_{0\to t},\widehat{I}_{1\to t})),
\end{aligned}
\end{equation}
\begin{equation}
\begin{aligned}
L_{cen} &= \psi(I_{t}^{GT},\widehat{I}_{0\to t}) + \psi(I_{t}^{GT},\widehat{I}_{1\to t})\\
&+\psi(I_{1},\Phi_{b}(O_{1\to t},\widehat{I}_{0\to t})) \\
&+ \psi(I_{0},\Phi_{b}(O_{0\to t},\widehat{I}_{1\to t})),
\end{aligned}
\end{equation}
where $\varphi(x) = \sqrt{x^2+\epsilon^2}$ is the Charbonnier function~\cite{lai2017deep}. The parameter $\epsilon$ is set to $1\times 10^{-6}$. $\psi(x)$ is the census function~\cite{meister2018unflow,zou2018df}, which is defined as the soft Hamming distance between census transformed image patches of size $7 \times 7$. $\widehat{I}_{0\to t} = \Phi_{b}(\widetilde{O}_{t\to0},I_{0})$ and $\widehat{I}_{1\to t} = \Phi_{b}(\widetilde{O}_{t\to1},I_{1})$ indicate the warped frames. $O_{1 \to t} = (1-t)\cdot O_{1 \to 0}$ and $O_{0 \to t} = t\cdot O_{0 \to 1}$ denote the optical flow to warp the $\widehat{I}_{0\to t}$ and $\widehat{I}_{1\to t}$, respectively.

They ensures the consistency between the consecutive frames. 
Finally, the total photometric loss $L_{pho}$ of this part is expressed as:
\begin{equation}
L_{pho} = L_{con} + L_{cen}.
\end{equation}
We use the Adamax optimizer~\cite{kingma2014adam} with $\beta_{1}=0.9$ and $\beta_{2}=0.999$, and use the batch size of $4$ for $20$ epochs. The initial learning rate is set as $5 \times 10^{-5}$ and then reduce the learning rate by a factor of $0.2$ when the losses of the testing set last for $4$ epochs without decreasing.

\subsubsection{Stage two}
In the second stage, we define the reconstruction loss $L_{rec}$ between the ground truth $I^{GT}_{t}$ and synthesized frame $I_{t}$ to train the entire model, it is defined as:
\begin{equation}
L_{rec} = \varphi(I^{GT}_{t}-I_{t}).
\end{equation}
Same as stage one, we use the same optimizer and learning rate reduction strategy. The initial learning rates of the consistent motion learning component and the trajectory-aware attention are set as $5 \times 10^{-5}$ and $5 \times 10^{-4}$, respectively. We jointly train the entire model for $70$ epochs. We also use the same strategies for reducing the learning rate and data augmentation as in the stage one.

\section{Experiments}
\label{e}
\subsection{Datasets and Metrics}

\subsubsection{Training dataset}
For fair comparisons, we follow existing works~\cite{gui2020featureflow,park2020bmbc,park2021asymmetric} to adopt a widely-used \textbf{Vimeo-90K} training set~\cite{xue2019video} to train our model. It has 51,312 triplets for training, where each triplet contains 3 consecutive video frames with a resolution of $256 \times 448$ pixels. We follow previous works~\cite{niklaus2018context,park2020bmbc,park2021asymmetric} to predict the middle frame and perform data augmentation by cropping $256\times256$ patches, flipping horizontally, flipping vertically, and reversing the temporal order of the triplet.

\begin{table*}[!t]
\renewcommand{\arraystretch}{1.1}
\begin{center}
\caption{Quantitative comparison (PSNR$\uparrow$ and SSIM$\uparrow$) on the Vimeo-90K~\cite{xue2019video}, UCF101~\cite{soomro2012ucf101} and DAVIS~\cite{perazzi2016benchmark} datasets. \textcolor{red}{Red} indicates the best and \underline{\textcolor{blue}{blue}} indicates the second best performance (best view in color).}
\label{table:1}
\begin{tabular}{l | c | c | c | c | c | c | c | c }
\hline

\hline
\multirow{2}*{Method}  &  \multirow{2}*{Runtime\ (seconds)} &  \multirow{2}*{\#Param\ (million)} &\multicolumn{2}{c|}{Vimeo-90K} & \multicolumn{2}{c|}{UCF101} & \multicolumn{2}{c}{DAVIS} \\
\cline{4-9} 
~ &  ~ &  ~ & PSNR(dB)$\uparrow$ & SSIM$\uparrow$ & PSNR(dB)$\uparrow$ & SSIM$\uparrow$ & PSNR(dB)$\uparrow$ & SSIM$\uparrow$ \\
\hline
ToFlow~\cite{xue2019video}        & 0.43 & 1.1   & 33.73 & 0.9682& 34.58 & 0.9667 &   25.39 & 0.8555  \\
SepConv~\cite{niklaus2017video2}   & 0.20 & 21.6 & 33.79 & 0.9702 & 34.78 & 0.9669 & 26.26 & 0.8610 \\
CAIN~\cite{choi2020channel}       & 0.04 & 42.8  & 34.65 & 0.9730& 34.91 & 0.9690 & 27.21 & 0.8730 \\
MEMC~\cite{bao2019memc}           & 0.12 & 70.3  & 34.29 & 0.9739& 34.96 & 0.9682 & 27.25 & 0.8914  \\
DAIN~\cite{bao2019depth}          & 0.13 & 24.0 & 34.70 & 0.9755 & 34.99 & 0.9683 & 27.31 & 0.8932 \\
AdaCoF~\cite{lee2020adacof}       & 0.03 & 22.9 & 34.35 & 0.9714& 35.16 & 0.9680  & 26.59 & 0.8707 \\
BMBC~\cite{park2020bmbc}          & 0.77 & 11.0 & 35.06 & 0.9766& 35.15 & 0.9688  & 26.95 & 0.8872 \\
EDSC~\cite{cheng2021multiple}     & 0.07 & 8.9  & 34.84 & 0.9750& 35.13 & 0.9680  &  26.99 & 0.8840  \\
ABME~\cite{park2021asymmetric}    & 0.22 & 18.1 & \underline{\textcolor{blue}{36.18}} & \underline{\textcolor{blue}{0.9805}} & \underline{\textcolor{blue}{35.38}} & \underline{\textcolor{blue}{0.9698}} & \underline{\textcolor{blue}{28.07}} & \underline{\textcolor{blue}{0.8984}} \\
\hline
\textbf{TTVFI}   & 0.35    &  16.6     & \textcolor{red}{36.54} & \textcolor{red}{0.9819}  & \textcolor{red}{35.51} & \textcolor{red}{0.9713}           &\textcolor{red}{28.31} & \textcolor{red}{0.9049} \\
\hline

\hline
\end{tabular}
\end{center}
\end{table*}

\begin{table*}[!t]
\renewcommand{\arraystretch}{1.1}
\begin{center}
\caption{Quantitative comparison (PSNR$\uparrow$ and SSIM$\uparrow$) on the SNU-FILE~\cite{choi2020channel} dataset. \textcolor{red}{Red} indicates the best and \underline{\textcolor{blue}{blue}} indicates the second best performance (best view in color).}
\label{table:2}
\begin{tabular}{l | c | c | c | c | c | c | c | c }
\hline

\hline
\multirow{2}*{Method} & \multicolumn{2}{c|}{Easy}  &\multicolumn{2}{c|}{Medium} & \multicolumn{2}{c|}{Hard} & \multicolumn{2}{c}{Extreme} \\
\cline{2-9} 
~ &  PSNR(dB)$\uparrow$ & SSIM$\uparrow$ & PSNR(dB)$\uparrow$ & SSIM$\uparrow$ & PSNR(dB)$\uparrow$ & SSIM$\uparrow$ & PSNR(dB)$\uparrow$ & SSIM$\uparrow$ \\
\hline
ToFlow~\cite{xue2019video}        & 39.08 & 0.9890  & 34.39 & 0.9740 & 28.44 & 0.9180 & 23.39 & 0.8310 \\
SepConv~\cite{niklaus2017video2}   & 39.41 & 0.9900  & 34.97 & 0.9762 & 29.36 & 0.9253 & 24.31 & 0.8448 \\
CAIN~\cite{choi2020channel}       & 39.89 & 0.9900  & 35.61 & 0.9776 & 29.90 & 0.9292 & 24.78 & 0.8507 \\
MEMC~\cite{bao2019memc}       & 39.92 & 0.9904  & 35.39 & 0.9779 & 29.93 & 0.9323 & 24.91 & 0.8561 \\
DAIN~\cite{bao2019depth}          & 39.73 & 0.9902  & 35.46 & 0.9780 & 30.17 & 0.9335 & 25.09 & 0.8584 \\
AdaCoF~\cite{lee2020adacof}       & 39.80 & 0.9900  & 35.05 & 0.9754 & 29.46 & 0.9244 & 24.31 & 0.8439 \\
BMBC~\cite{park2020bmbc}          & 39.90 & \underline{\textcolor{blue}{0.9902}}  & 35.31 & 0.9774 & 29.33 & 0.9270 & 23.92 & 0.8432 \\
EDSC~\cite{cheng2021multiple}     & \underline{\textcolor{blue}{40.01}} & 0.9900  & 35.37 & 0.9780 & 29.59 & 0.9260 & 24.39 & 0.8430 \\
ABME~\cite{park2021asymmetric}    & 39.59 & 0.9901  & \underline{\textcolor{blue}{35.77}} & \underline{\textcolor{blue}{0.9789}} & \underline{\textcolor{blue}{30.58}} & \underline{\textcolor{blue}{0.9364}} & \underline{\textcolor{blue}{25.42}} & \underline{\textcolor{blue}{0.8639}} \\
\hline
\textbf{TTVFI}   & \textcolor{red}{40.22} & \textcolor{red}{0.9907}   & \textcolor{red}{36.07} & \textcolor{red}{0.9794}   & \textcolor{red}{30.77} & \textcolor{red}{0.9397}     & \textcolor{red}{25.67} & \textcolor{red}{0.8743}     \\
\hline

\hline
\end{tabular}
\end{center}
\end{table*}

\subsubsection{Test datasets}
We evaluate the proposed TTVFI and compare its performance with other SOTA approaches on four widely used test sets: {Vimeo-90K}~\cite{xue2019video}, {UCF101}~\cite{liu2017video}, {DAVIS}~\cite{perazzi2016benchmark}, and {SNU-FILM}~\cite{choi2020channel}. 
\\[4pt]
\textbf{Vimeo-90K} is the Vimeo-90K testing set~\cite{xue2019video} and contains 3,782 triplets of spatial resolution $256\times448$. 
\\[4pt]
\textbf{UCF101} is the constructed test set by selecting from the human action videos dataset UCF101~\cite{soomro2012ucf101} and contains 379 triplets of spatial resolution $256\times256$. 
\\[4pt]
\textbf{DAVIS} is the constructed test set by selecting from the video object segmentation dataset DAVIS~\cite{perazzi2016benchmark} and contains 30 triplets of different spatial resolutions.
\\[4pt]
\textbf{SNU-FILM} contains a total of 1,240 triplets videos, depending on the complexity of the motion, it has four different settings–Easy, Medium, Hard, and Extreme. Each part contains 310 triplets videos with a resolution of $1280\times720$. 

\subsubsection{Evaluation metrics}
For fair comparisons, we follow previous works~\cite{bao2019depth,cheng2021multiple,park2020bmbc,park2021asymmetric} to use peak signal-to-noise ratio (PSNR) and structural similarity index (SSIM)~\cite{wang2004image} as a widely used metric for evaluating.

\subsection{Comparisons with State-of-the-art Methods}
We compare TTVFI with nine classical start-of-the-art methods. 
These methods can be summarized into three categories: CNN-based~\cite{choi2020channel}, kernel-based~\cite{cheng2021multiple,lee2020adacof,niklaus2017video2}, and flow-based video interpolation~\cite{bao2019depth,bao2019memc,park2020bmbc,park2021asymmetric,xue2019video}. 
For fair comparisons, we obtain the performance from their original paper or reproduce results by authors' officially released models.

\subsubsection{Quantitative comparison}
As shown in Tab.~\ref{table:1}, the results for each algorithm on the three test sets: Vimeo-90K~\cite{xue2019video}, UCF101~\cite{soomro2012ucf101}, and DAVIS~\cite{perazzi2016benchmark}. 
Benefiting from a pure CNN structure, CAIN~\cite{choi2020channel} uses less inference time, but it does not handle motion well and has poor performance. Although the kernel-based methods (e.g.,~AdaCoF~\cite{lee2020adacof}, EDSC~\cite{cheng2021multiple}) achieve better performance than CAIN~\cite{choi2020channel}, the kernel size directly restricts the motion that the model can capture, resulting in heavy memory and computation cost. Thanks to the progress of motion estimation, the latest flow-based methods (e.g.,~ABME~\cite{park2021asymmetric}, BMBC~\cite{park2020bmbc}) generally perform better than the kernel-based methods. However, under some challenging conditions that decrease the accuracy of optical flow, these methods only blend warped frames through the synthesis network can lead to suboptimal performance. 

TTVFI introduces more pristine features of the intermediate frame from original input frames by motion trajectories. It achieves a result of 36.54dB, 35.51dB, and 28.31dB PSNR and significantly outperforms the other algorithms for all test sets by a large margin. Specifically, on the Vimeo-90K~\cite{xue2019video} and DAVIS~\cite{perazzi2016benchmark} datasets, TTVFI outperforms ABME~\cite{park2021asymmetric} by \textbf{0.36dB} and \textbf{0.24dB}, respectively. This large margin demonstrates the power of TTVFI in feature restoration. Besides, we follow previous works~\cite{bao2019depth,park2020bmbc,park2021asymmetric} to report the runtime of interpolating a frame of size $640\times480$ by using an RTX 2080 Ti GPU. TTVFI achieves higher performance while keeping the comparable Runtime and \#Params.

\begin{figure*}[!t]
\centering
\includegraphics[width=1.0\linewidth]{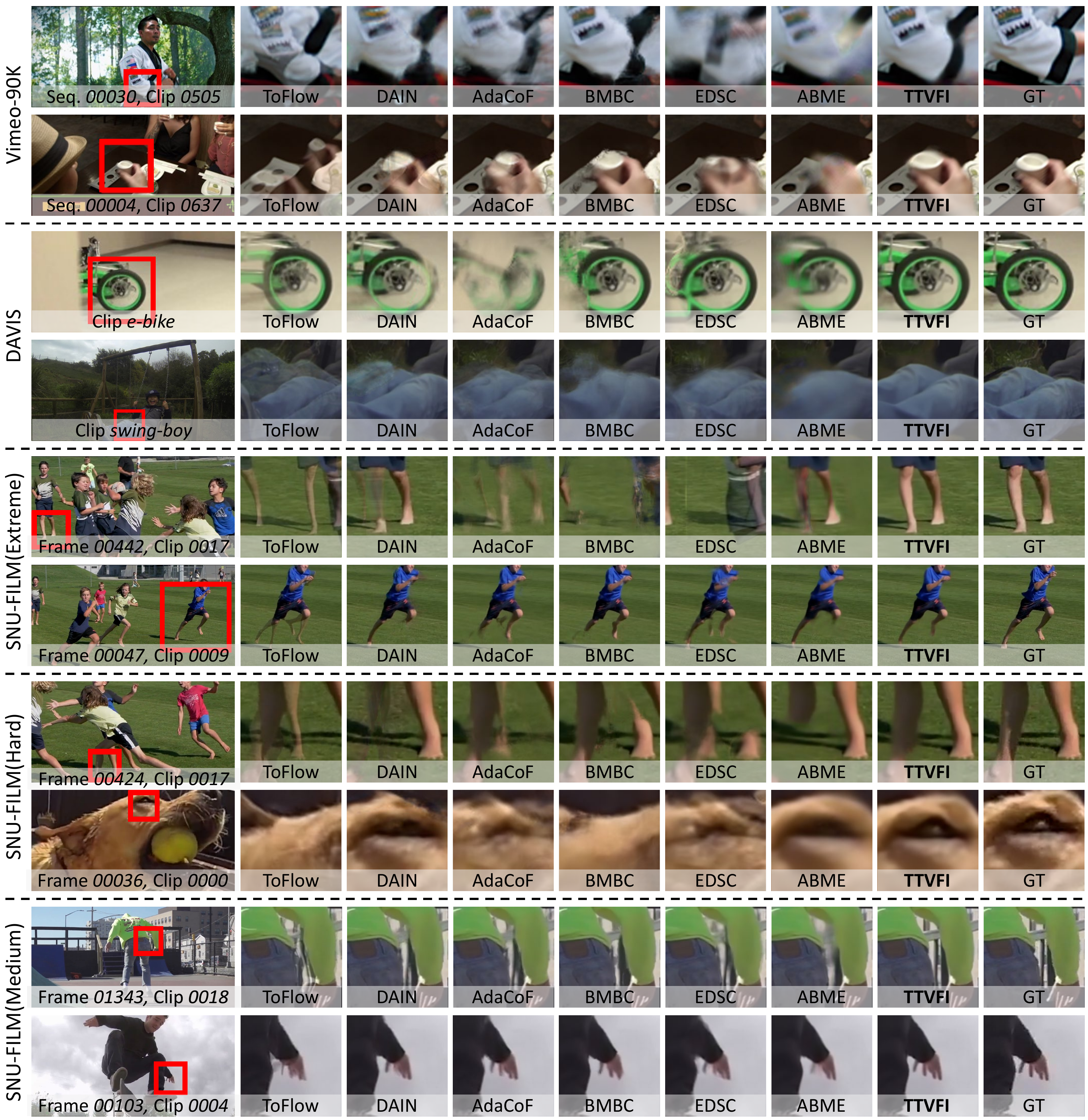}
\caption{Visual results on Vimeo-90K~\cite{xue2019video}, DAVIS~\cite{perazzi2016benchmark}, and SNU-FILM~\cite{choi2020channel} datasets. The frame number and method are shown at the bottom of each case. Zoom in to see better visualization.}
\label{fig:case}
\end{figure*}

To further verify the generalization capabilities of TTVFI, we evaluate TTVFI on SNU-FILE~\cite{choi2020channel} dataset with different complexities. As shown in Tab.~\ref{table:2}, due to the well-designed tokens and the long-range modeling capability of attention mechanism, TTVFI achieves better results in all four kinds of settings, which outperforms other SOTA methods between \textbf{0.19dB} to \textbf{0.30dB}. The performances verify that TTVFI has strong generalization capabilities under different degrees of motion. More results for perceptual metrics can be found in the supplementary material.

\subsubsection{Qualitative comparison}
To further compare the visual qualities of different approaches, we show visual results generated by TTVFI and other SOTA methods on different test sets in Fig.~\ref{fig:case}. For fair comparisons, we either directly take the original interpolated results of the author-released or use author-released models to get results. It can be observed that TTVFI has a great improvement in visual quality, especially for areas with moving instances. For example, in the fifth row in Fig.~\ref{fig:case}, TTVFI can recover the complete leg in the case of extreme motion. As the analysis mentioned above, the results verify that TTVFI can mitigate the distortion and blur caused by inconsistent warping. More visual results can be found in the supplementary materials. 

\begin{table}[!t]
\begin{center}
\caption{Ablation study of our TTVFI on Vimeo-90K~\cite{xue2019video} and DAVIS~\cite{perazzi2016benchmark} datasets. CML: consistent motion learning module. TAC: trajectory-aware attention with consistent tokens. TAB: trajectory-aware attention with boundary tokens.}
\label{table:3}
\begin{tabular}{cccc | c | c | c | c}
\hline

\hline
\multicolumn{4}{c|}{\textbf{Components}} &  \multicolumn{2}{c|}{Vimeo-90K} & \multicolumn{2}{c}{DAVIS} \\
\cline{5-8}
Base   &         CML &     TAC &   TAB  & PSNR& SSIM & PSNR& SSIM  \\
\hline
\checkmark &       ~     & ~          & ~         & 34.26& 0.9724        & 27.28& 0.8939  \\
\checkmark &  \checkmark & ~          & ~         & 34.95& 0.9755        & 27.70& 0.8981  \\
\checkmark & \checkmark  & \checkmark & ~         & 36.45& 0.9815        & 28.21& 0.9025  \\
\checkmark & \checkmark  & ~          & \checkmark         & 36.31& 0.9811        & 28.11& 0.9003  \\
\checkmark & \checkmark  & \checkmark & \checkmark& \textbf{36.54}& \textbf{0.9819}        & \textbf{28.31}& \textbf{0.9049} \\
\hline

\hline
\end{tabular}
\end{center}
\end{table}

\begin{figure*}[!t]
\centering
\includegraphics[width=0.9\linewidth]{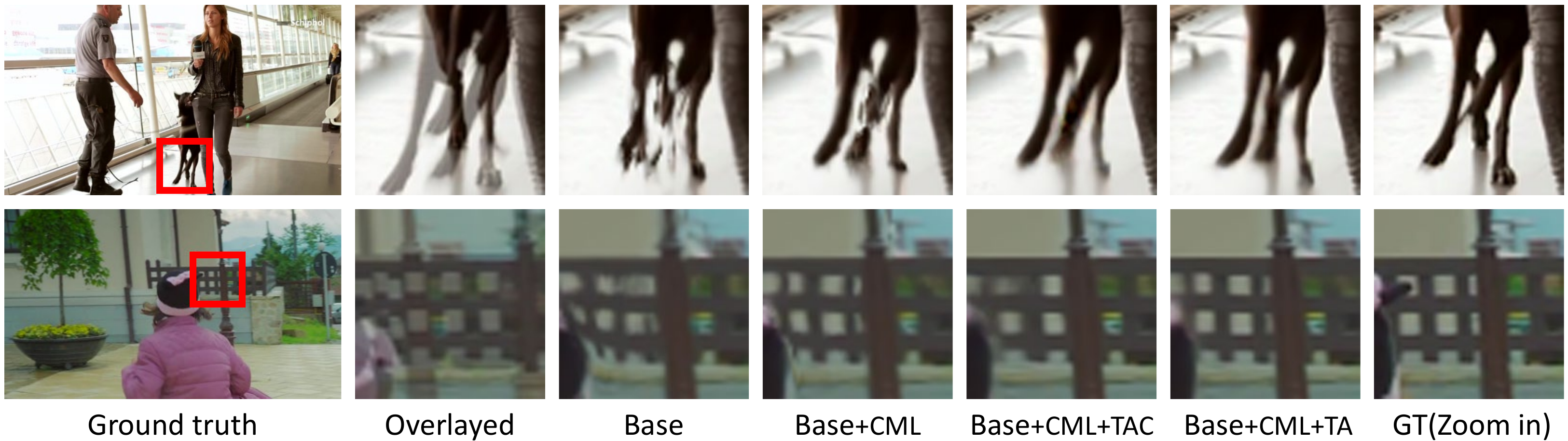}
\caption{Ablation study on the consistent motion learning module (CML), trajectory-aware attention with consistent tokens (TAC), and trajectory-aware attention (TA). Zoom in to see better visualization. (``TA'' can be interpreted as ``TAC+TAB'').}
\label{fig:ablation}
\end{figure*}

\begin{figure*}[!t]
\centering
\includegraphics[width=0.9\linewidth]{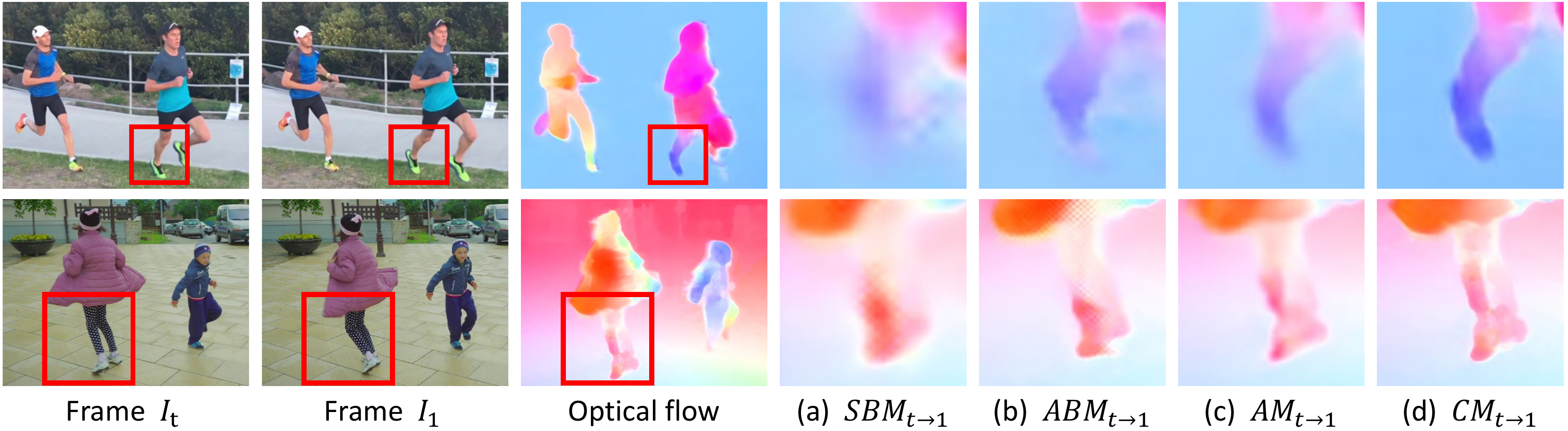}
\caption{Visualization comparison of different motion fields. (a) symmetric bilateral motion in BMBC~\cite{park2020bmbc}, (b) asymmetric bilateral motion in ABME~\cite{park2021asymmetric}, (c) approximated motion, (d) consistent motion. Zoom in to see better visualization.}
\label{fig:motion}
\end{figure*}

\subsection{Ablation Study}
In this section, we mainly conduct the ablation study on the proposed trajectory-aware Transformer and analyze the effect of the used motion field in consistent motion learning module.

\subsubsection{Trajectory-aware Transformer}
Our proposed trajectory-aware Transformer includes three important parts, ``CML", ``TAC", and ``TAB". ``CML" is the consistent motion learning module to generate the trajectory, ``TA" is the trajectory-aware attention. Depending on the consistent tokens and boundary tokens being used, ``TA" can be divided into ``TAC" and TAB". Trajectory-aware Transformer can be interpreted as ``Base+CML+TA(TAC+TAB)", and we study them together in this part. Among them, we directly use queries to generate the feature of the intermediate frame as our ``Base" model. The ``Base+CML" denotes that we further aggregate the features warped by consistent motion. Then we add the trajectory-aware attention based on the consistent tokens and boundary tokens as our ``Base+CML+TAC" and ``Base+CML+TAB" model, respectively. We add the trajectory-aware attention based on the both tokens progressively as our ``Base+CML+TAC+TAB" model. 

As shown in Tab.~\ref{table:3}, the addition of CML improves the PSNR from 34.26dB to 34.95dB on Vimeo-90K~\cite{xue2019video} and from 27.28dB to 27.70dB on DAVIS~\cite{perazzi2016benchmark} dataset. With the addition of TAC, the performance is improved from {34.95dB} to {36.45dB} on Vimeo-90K~\cite{xue2019video} and from {27.70dB} to {28.21dB} on DAVIS~\cite{perazzi2016benchmark}, respectively. With the addition of TAB, the performance is improved from {34.95dB} to {36.31dB} on Vimeo-90K~\cite{xue2019video} and from {27.70dB} to {28.11dB} on DAVIS~\cite{perazzi2016benchmark}, respectively. With all of them added (i.e.,~Base+CML+TA(TAC+TAB)), the performance has achieved 36.54dB and 28.31dB on Vimeo-90K~\cite{xue2019video} and DAVIS~\cite{perazzi2016benchmark}, respectively. This demonstrates the superiority of each part in TTVFI.

\begin{table}[!t]
\begin{center}
\caption{Results of using different motion fields on Vimeo-90K~\cite{xue2019video} and DAVIS~\cite{perazzi2016benchmark} datasets.}
\label{table:5}
\begin{tabular}{cc | c | c | c | c} 
\hline

\hline
\multicolumn{2}{c|}{\textbf{Motion field}} &  \multicolumn{2}{c|}{Vimeo-90K} & \multicolumn{2}{c}{DAVIS} \\
\cline{3-6}
Approximation   &  Consistent   & PSNR&SSIM    & PSNR&SSIM  \\
\hline
\checkmark &       ~     & 35.80&0.9795     & 27.72&0.8974  \\
      ~    &  \checkmark & 35.87&0.9799      & 27.91&0.8997           \\
\checkmark & \checkmark  & \textbf{36.54}&\textbf{0.9819} & \textbf{28.31}&\textbf{0.9049}  \\
\hline

\hline
\end{tabular}
\end{center}
\end{table}

\begin{figure*}[!t]
\centering
\includegraphics[width=0.85\linewidth]{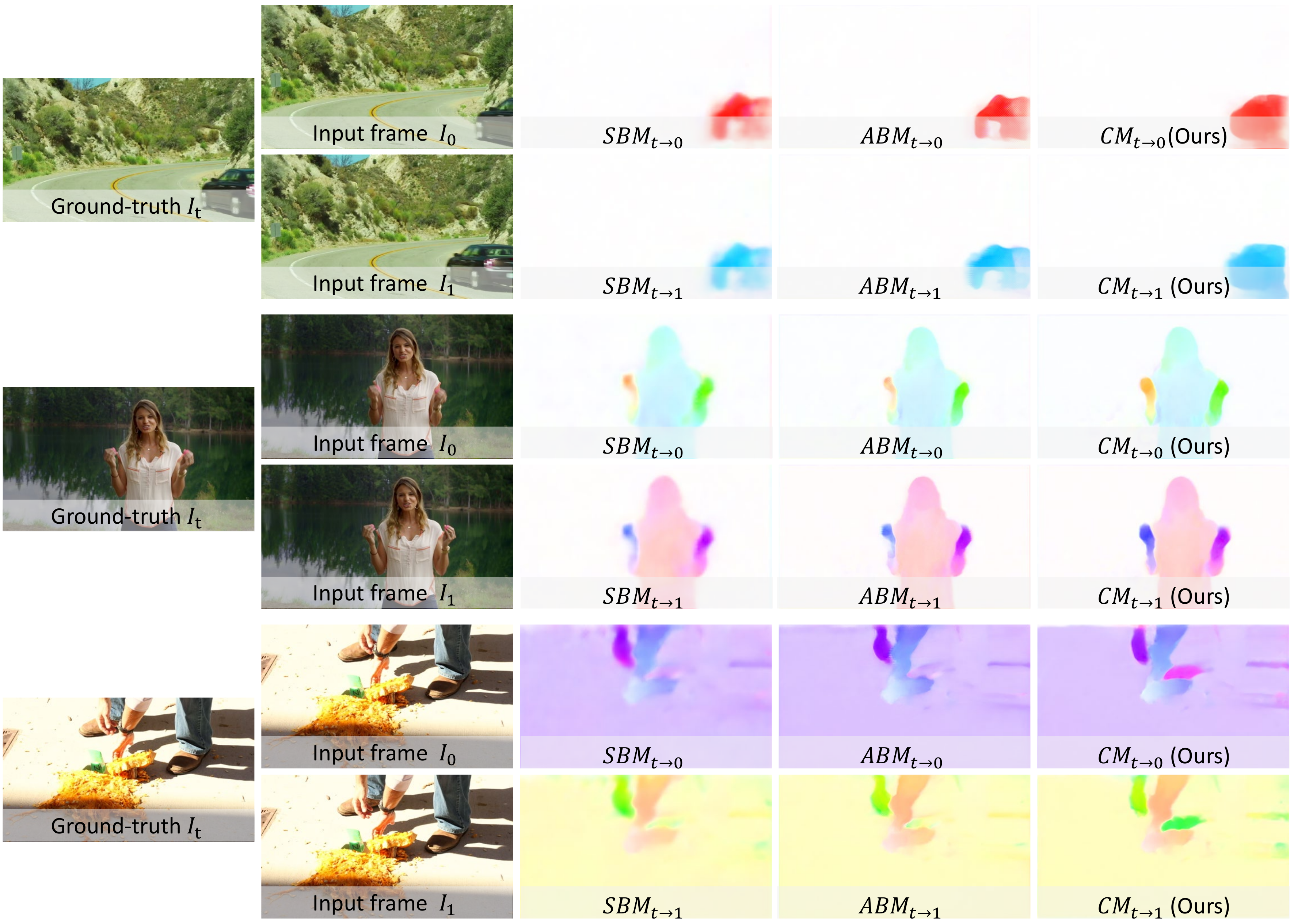}
\caption{Visualization comparison of consistent motion $CM$ with symmetric bilateral motion $SBM$ (i.e.,~BMBC~\cite{park2020bmbc}) and asymmetric bilateral motion $ABM$ (i.e.,~ABME~\cite{park2021asymmetric}).}
\label{fig:smsf}
\end{figure*}

To further compare the visual qualities of different approaches, we further compare them as shown in Fig.~\ref{fig:ablation}. For fair comparisons, we use the same experimental setup for the following comparison. It can be observed that each part of the TTVFI has a significant contribution to improving visual quality. For example, in the second row in Fig.~\ref{fig:ablation}, CML can improve the accuracy of motion fields, while TAC and TAB integrate tokens into the features to produce more complete and clearer fence structures.

\begin{table}[!t]
\begin{center}
\caption{Results of trajectory-aware attention with different multi-head ($H$), window size ($S$), and layer number ($N$) on Vimeo-90K~\cite{xue2019video} dataset.}
\label{table:4}
\begin{tabular}{c |c || c | c|| c | c}
\hline

\hline
$H$ &  PSNR/SSIM & $S$ &  PSNR/SSIM & $N$ &  PSNR/SSIM \\
\hline
2         & 36.52/0.9818 & 4          & 36.43/0.9814   &1          & 36.31/0.9812  \\
4          & 36.54/0.9819 & 8         & 36.54/0.9819     & 2          & 36.54/0.9819         \\
8          & 36.53/0.9819 & 12          & 36.50/0.9818    & 3          & 36.58/0.9820         \\
\hline

\hline
\end{tabular}
\end{center}
\end{table}

\subsubsection{Motion field in consistent motion learning module}

To verify the effectiveness of consistent motion generated by the consistent motion learning component.
We chose different motions that described in Sec.~\ref{tg} in our method to perform attention. As shown in Tab.~\ref{table:5}, using the consistent motion is better than approximated motion. With all of them added, the performance has achieved 36.54dB and 28.31dB on Vimeo-90K~\cite{xue2019video} and DAVIS~\cite{perazzi2016benchmark}, respectively. This demonstrates the superiority of consistent motion learning component in TTVFI. 

To further compare the visual qualities of different motion field used, we further compare the visual differences of them as shown in Fig.~\ref{fig:motion}. The consistent motion (i.e.,~$CM$) has clearer textures than other kinds of motion fields. Both quantitative and qualitative comparisons demonstrate the superiority of consistent motion.

\begin{table}[!t]
\begin{center}
\caption{Results of stacked trajectory-aware Transfomer on multi-scales on Vimeo-90K~\cite{xue2019video} dataset.}
\label{table:6}
\begin{tabular}{ccc | c | c } 
\hline

\hline
\multicolumn{3}{c|}{\textbf{Scale factor}} &  \multirow{2}*{PSNR} & \multirow{2}*{SSIM} \\
$\times1$   &  $\times2$   & $\times4$    & ~  & ~   \\
\hline
\checkmark &       ~      &       ~    & 36.06  & 0.9762  \\
\checkmark &  \checkmark &        ~  &  36.31 & 0.9806  \\
\checkmark & \checkmark  & \checkmark  & 36.54 & 0.9819  \\
\hline

\hline
\end{tabular}
\end{center}
\end{table}

\section{Discussions}
\label{d}
In this section, we mainly discuss the influence of hyper-parameters used in the attention mechanism, different motion field, multi-scale fusion structure, and inconsistent region map.

\begin{figure*}[!t]
\centering
\includegraphics[width=0.85\linewidth]{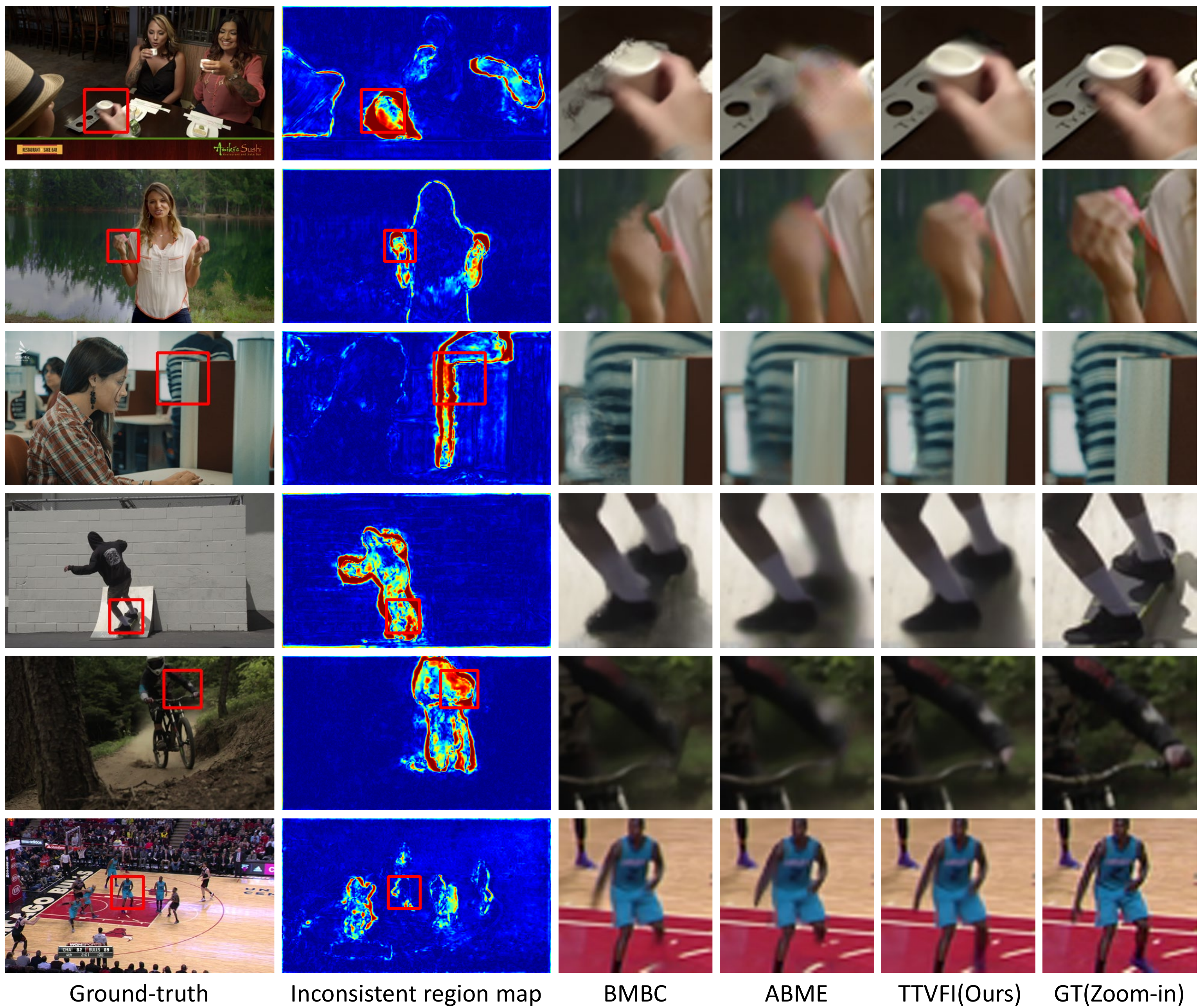}
\caption{Visualization of inconsistent region map $\mathcal{P}$, and results comparison with BMBC~\cite{park2020bmbc} and ABME~\cite{park2021asymmetric}. Zoom in to see better visualization.}
\label{fig:sirm}
\end{figure*}

\subsection{The Discussions of Hyper-parameters in Trajectory-aware Attention}

To explore the influence of hyper-parameters used in attention mechanisms that described in Sec.~\ref{ja}. We discuss the different multi-head ($H$), window size ($S$), and layer number ($N$) in attention mechanisms, as shown in Tab.~\ref{table:4}. The impact of $H$ is insignificant since the dimension of features is small. Proper $S$ can effectively model spatial motion without introducing useless or insufficient information. The performance is positively correlated with the $N$, it demonstrates the learning ability of the trajectory-aware attention. However, a deeper hierarchical structure with limited improvements will introduce heavy memory and computation cost. 
After a trade-off between performance improvement and computational cost growth, we choose 4, 8, 2 as the value of $H$, $S$, and $N$.

\begin{table}[!t]
\centering
\caption{Results of generating inconsistent regions with different temperature coefficients $\tau$ on Vimeo-90K~\cite{xue2019video} dataset.}
\label{table:7}
\begin{tabular}{c|c|c|c}
    \hline
    
    \hline
    \textbf{$\tau$}     &   0.5  &  1.0  &  2.0  \\
    \hline
    PSNR  & 36.51 & 36.54 & 36.54 \\
    \hline
    SSIM  & 0.9818 & 0.9819 & 0.9819 \\
    \hline
    
    \hline
\end{tabular}
\end{table}

\begin{figure}[!t]
\centering
\includegraphics[width=1.0\linewidth]{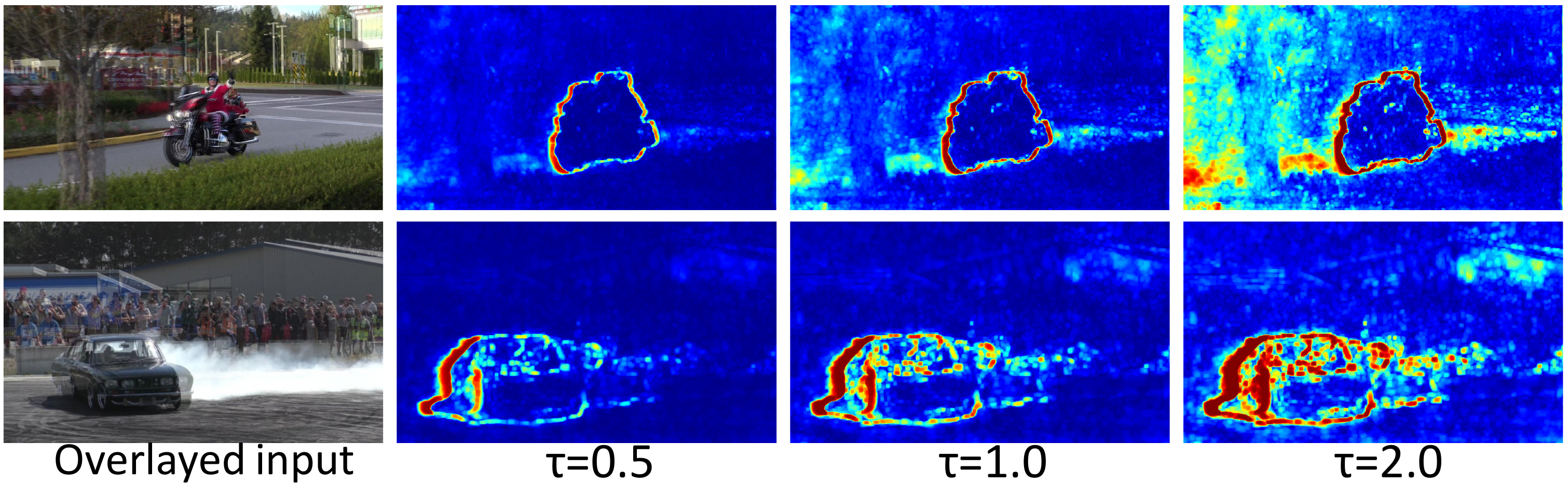}
\caption{Visualization comparison of inconsistent region map with different temperature coefficients $\tau$ in sigmoid function. Zoom in to see better visualization.}
\label{fig:conf}
\end{figure}

\subsection{The Discussion of Different Motion Field}

To verify the effectiveness of consistent motion generated by the consistent motion learning component that described in Sec.~\ref{cmlc}. As shown in Fig.~\ref{fig:smsf}, we compare the visual qualities of consistent motion with other state-of-the-art flow-based algorithms, such as BMBC~\cite{park2020bmbc} and ABME~\cite{park2021asymmetric}. The consistent motion has clearer textures, which indicates the superiority of the generated consistent motion field.

\begin{figure*}[!t]
  \centering
  \includegraphics[width=0.7\linewidth]{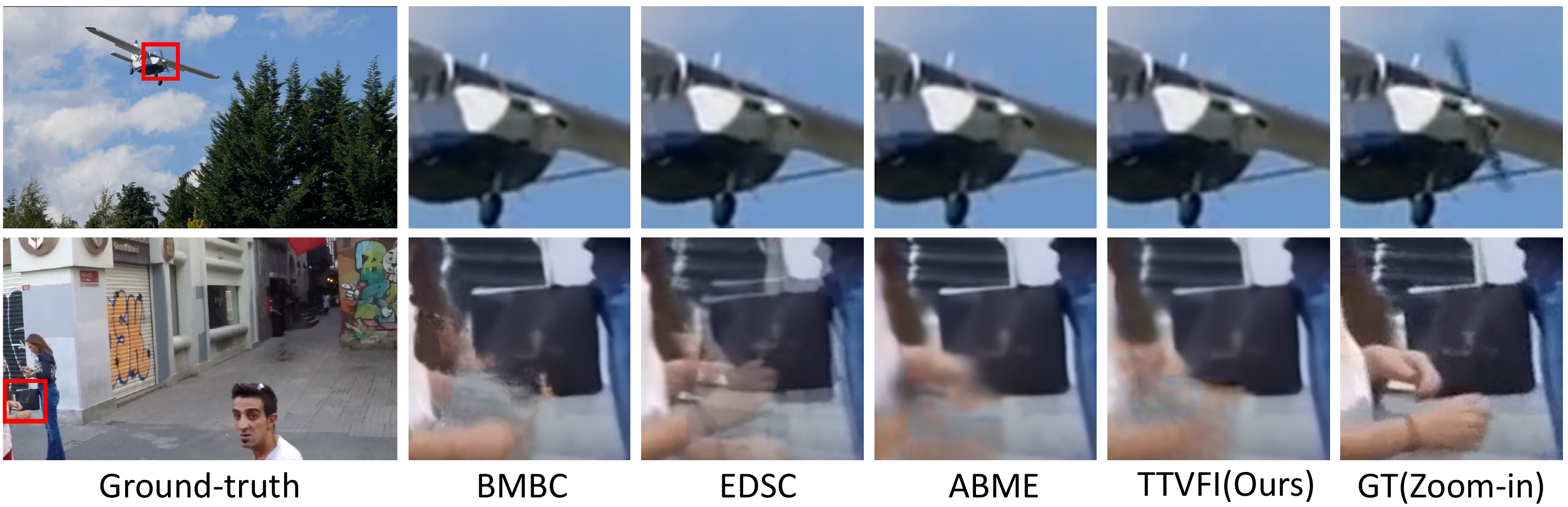}
  \caption{Failure case when rotation and camera motion occur, and the results comparison with other SOTA methods, such as BMBC~\cite{park2020bmbc}, EDSC~\cite{cheng2021multiple}, and ABME~\cite{park2021asymmetric}.}
  \label{fig:FC}
\end{figure*}

\subsection{The Discussion of Multi-scale Fusion Structure}

In the previous works~\cite{yang2020learning,liu2021swin}, stacking transformers in multi-layer and multi-scale has been proven to be effective. Therefore, as described in Sec.~\ref{msf} and shown in Fig.~\ref{fig:msf}, we stack the trajectory-aware Transformer at multi-scale (i.e. $\times1$, $\times2$, and $\times4$), and each scale contains multiple layers of attention mechanisms. In this section, we investigate the effects of Transformer multi-scale stacking on performance. As shown in Tab.~\ref{table:6}, the features of multi-scale can effectively facilitate the interaction of multi-scale features and improve the performance. This demonstrates the effectiveness of feature fusion and interaction in multi-scale.

\subsection{The Discussion of Inconsistent Region Map}
Our method targets the inconsistent motion regions that are the source of distortion and blur. The inconsistent region map is used to distinguish the different regions of inconsistent motion. To verify the effectiveness of inconsistent region map $\mathcal{P}$ in attention calculation, as described in Sec.~\ref{ja}, we visualize it and compare the interpolation results with other methods. As shown in Fig.~\ref{fig:sirm}, based on the inconsistent region map, our method introduces pristine features of original input frames into the intermediate frame along the trajectories and gets better results.

Besides, the sigmoid function is used to generate the inconsistent motion map, so we explore the sensitivity of the temperature coefficient $\tau$ in the sigmoid function~\footnote{$\text{Sigmoid}(x)=\frac{1}{1+e^{-\tau\cdot x}}$} and visualize the inconsistent region map. As shown in Fig.~\ref{fig:conf}, regions of inconsistent motion are mainly concentrated at the boundary of moving instances, and the area of the inconsistent motion increases with increasing $\tau$. Notably, as shown in Tab.~\ref{table:7}, $\tau$ is insensitive to the final performance. It is insensitive to the final performance, which indicates that our method of generating the inconsistent motion region is reliable and robust enough.

\section{Limitations}
\label{l}
In this section, we discuss the limitations of TTVFI and the failure cases as shown in Fig.~\ref{fig:FC}.

\subsubsection{Rotation} Although we propose a consistent motion learning component to generate consistent motion, when facing complex motion (e.g., rotation), as shown in the upper part of Fig.~\ref{fig:FC}, the accuracy of the motion trajectory is limited and the significance is reduced. 

\subsubsection{Camera motion} We propose to pay more attention to the regions with inconsistent motion, which usually focuses on the moving instances in the frame. However, when intense camera motion occurs, the motion of frame boundaries is inconsistent and incomplete. As shown in the bottom part of Fig.~\ref{fig:FC}, the incomplete motion makes little information that can be used to recover frames boundaries, leading to poor results.

\section{Conclusions}
\label{c}
In this paper, we pay more attention to the important synthesis network in VFI and propose a new trajectory-aware transformer (TTVFI). In particular, TTVFI aims to mitigate the distortion and blur caused by inconsistent motion and inaccurate warping in existing algorithms, and learns more accurate features of the intermediate frames from the original input frames.
To implement such formulations better, we first propose a consistent motion learning component to generate the consistent motion field, which can be defined as a group of inter-frame motion trajectories. Then we formulate video frames into two kinds of pre-aligned visual tokens and calculate attention separately according to whether the regional motion is consistent or not. 
To our best knowledge, TTVFI is the first work to enables Transformers to model the features of intermediate frames by motion trajectory in VFI. Experimental results show the superiority between the proposed TTVFI and existing SOTA methods. 

In the future, we will focus on 1) extending the inter-frame motion trajectories to more input frames in the VFI task, and 2) transferring the trajectory-aware Transformer in more low-level vision tasks by more explorations.

\bibliographystyle{plain}
\bibliography{ref}

\begin{thebibliography}{10}

\bibitem{bao2019depth}
Wenbo Bao, Wei-Sheng Lai, Chao Ma, Xiaoyun Zhang, Zhiyong Gao, and Ming-Hsuan
  Yang.
\newblock Depth-aware video frame interpolation.
\newblock In {\em CVPR}, pages 3703--3712, 2019.

\bibitem{bao2019memc}
Wenbo Bao, Wei-Sheng Lai, Xiaoyun Zhang, Zhiyong Gao, and Ming-Hsuan Yang.
\newblock {MEMC-Net}: Motion estimation and motion compensation driven neural
  network for video interpolation and enhancement.
\newblock {\em IEEE TPAMI}, 2019.

\bibitem{bao2018high}
Wenbo Bao, Xiaoyun Zhang, Li~Chen, Lianghui Ding, and Zhiyong Gao.
\newblock High-order model and dynamic filtering for frame rate up-conversion.
\newblock {\em IEEE TIP}, 27(8):3813--3826, 2018.

\bibitem{cao2021video}
Jiezhang Cao, Yawei Li, Kai Zhang, and Luc Van~Gool.
\newblock Video super-resolution transformer.
\newblock {\em arXiv preprint arXiv:2106.06847}, 2021.

\bibitem{carion2020end}
Nicolas Carion, Francisco Massa, Gabriel Synnaeve, Nicolas Usunier, Alexander
  Kirillov, and Sergey Zagoruyko.
\newblock End-to-end object detection with transformers.
\newblock In {\em ECCV}, pages 213--229, 2020.

\bibitem{cheng2020video}
Xianhang Cheng and Zhenzhong Chen.
\newblock Video frame interpolation via deformable separable convolution.
\newblock In {\em AAAI}, volume~34, pages 10607--10614, 2020.

\bibitem{cheng2021multiple}
Xianhang Cheng and Zhenzhong Chen.
\newblock Multiple video frame interpolation via enhanced deformable separable
  convolution.
\newblock {\em IEEE TPAMI}, 2021.

\bibitem{choi2020channel}
Myungsub Choi, Heewon Kim, Bohyung Han, Ning Xu, and Kyoung~Mu Lee.
\newblock Channel attention is all you need for video frame interpolation.
\newblock In {\em AAAI}, volume~34, pages 10663--10671, 2020.

\bibitem{dai2017deformable}
Jifeng Dai, Haozhi Qi, Yuwen Xiong, Yi~Li, Guodong Zhang, Han Hu, and Yichen
  Wei.
\newblock Deformable convolutional networks.
\newblock In {\em ICCV}, pages 764--773, 2017.

\bibitem{dosovitskiy2020image}
Alexey Dosovitskiy, Lucas Beyer, Alexander Kolesnikov, Dirk Weissenborn,
  Xiaohua Zhai, Thomas Unterthiner, Mostafa Dehghani, Matthias Minderer, Georg
  Heigold, Sylvain Gelly, et~al.
\newblock An image is worth 16x16 words: Transformers for image recognition at
  scale.
\newblock {\em arXiv preprint arXiv:2010.11929}, 2020.

\bibitem{flynn2016deepstereo}
John Flynn, Ivan Neulander, James Philbin, and Noah Snavely.
\newblock Deepstereo: Learning to predict new views from the world's imagery.
\newblock In {\em CVPR}, pages 5515--5524, 2016.

\bibitem{fourure2017gridnet}
Damien Fourure, R{\'e}mi Emonet, Elisa Fromont, Damien Muselet, Alain
  Tr{\'e}meau, and Christian Wolf.
\newblock Residual conv-deconv grid network for semantic segmentation.
\newblock In {\em BMVC}, pages 181.1--181.13, 2017.

\bibitem{gui2020featureflow}
Shurui Gui, Chaoyue Wang, Qihua Chen, and Dacheng Tao.
\newblock Feature{F}low: Robust video interpolation via structure-to-texture
  generation.
\newblock In {\em CVPR}, pages 14004--14013, 2020.

\bibitem{he2015delving}
Kaiming He, Xiangyu Zhang, Shaoqing Ren, and Jian Sun.
\newblock Delving deep into rectifiers: Surpassing human-level performance on
  imagenet classification.
\newblock In {\em ICCV}, pages 1026--1034, 2015.

\bibitem{jia2016dynamic}
Xu~Jia, Bert De~Brabandere, Tinne Tuytelaars, and Luc~V Gool.
\newblock Dynamic filter networks.
\newblock {\em NeurIPS}, 29, 2016.

\bibitem{jiang2018super}
Huaizu Jiang, Deqing Sun, Varun Jampani, Ming-Hsuan Yang, Erik Learned-Miller,
  and Jan Kautz.
\newblock Super slomo: High quality estimation of multiple intermediate frames
  for video interpolation.
\newblock In {\em CVPR}, pages 9000--9008, 2018.

\bibitem{kingma2014adam}
Diederik~P Kingma and Jimmy Ba.
\newblock Adam: A method for stochastic optimization.
\newblock In {\em ICLR}, 2015.

\bibitem{lai2017deep}
Wei-Sheng Lai, Jia-Bin Huang, Narendra Ahuja, and Ming-Hsuan Yang.
\newblock Deep laplacian pyramid networks for fast and accurate
  super-resolution.
\newblock In {\em CVPR}, pages 624--632, 2017.

\bibitem{lee2020adacof}
Hyeongmin Lee, Taeoh Kim, Tae-young Chung, Daehyun Pak, Yuseok Ban, and
  Sangyoun Lee.
\newblock {AdaCoF}: Adaptive collaboration of flows for video frame
  interpolation.
\newblock In {\em CVPR}, pages 5316--5325, 2020.

\bibitem{liu2022learning}
Chengxu Liu, Huan Yang, Jianlong Fu, and Xueming Qian.
\newblock Learning trajectory-aware transformer for video super-resolution.
\newblock In {\em CVPR}, pages 5687--5696, 2022.

\bibitem{liu2019deep}
Yu-Lun Liu, Yi-Tung Liao, Yen-Yu Lin, and Yung-Yu Chuang.
\newblock Deep video frame interpolation using cyclic frame generation.
\newblock In {\em AAAI}, volume~33, pages 8794--8802, 2019.

\bibitem{liu2021swin}
Ze~Liu, Yutong Lin, and et~al.
\newblock Swin {T}ransformer: Hierarchical vision transformer using shifted
  windows.
\newblock In {\em ICCV}, pages 10012--10022, 2021.

\bibitem{liu2017video2}
Ziwei Liu, Raymond~A Yeh, Xiaoou Tang, Yiming Liu, and Aseem Agarwala.
\newblock Video frame synthesis using deep voxel flow.
\newblock In {\em ICCV}, pages 4463--4471, 2017.

\bibitem{liu2017video}
Ziwei Liu, Raymond~A Yeh, Xiaoou Tang, Yiming Liu, and Aseem Agarwala.
\newblock Video frame synthesis using deep voxel flow.
\newblock In {\em ICCV}, pages 4463--4471, 2017.

\bibitem{meister2018unflow}
Simon Meister, Junhwa Hur, and Stefan Roth.
\newblock Un{F}low: Unsupervised learning of optical flow with a bidirectional
  census loss.
\newblock In {\em AAAI}, 2018.

\bibitem{niklaus2018context}
Simon Niklaus and Feng Liu.
\newblock Context-aware synthesis for video frame interpolation.
\newblock In {\em CVPR}, pages 1701--1710, 2018.

\bibitem{niklaus2020softmax}
Simon Niklaus and Feng Liu.
\newblock Softmax splatting for video frame interpolation.
\newblock In {\em CVPR}, pages 5437--5446, 2020.

\bibitem{niklaus2017video2}
Simon Niklaus, Long Mai, and Feng Liu.
\newblock Video frame interpolation via adaptive convolution.
\newblock In {\em CVPR}, pages 670--679, 2017.

\bibitem{niklaus2017video}
Simon Niklaus, Long Mai, and Feng Liu.
\newblock Video frame interpolation via adaptive separable convolution.
\newblock In {\em ICCV}, pages 261--270, 2017.

\bibitem{park2020bmbc}
Junheum Park, Keunsoo Ko, Chul Lee, and Chang-Su Kim.
\newblock {BMBC}: Bilateral motion estimation with bilateral cost volume for
  video interpolation.
\newblock In {\em ECCV}, pages 109--125. Springer, 2020.

\bibitem{park2021asymmetric}
Junheum Park, Chul Lee, and Chang-Su Kim.
\newblock Asymmetric bilateral motion estimation for video frame interpolation.
\newblock In {\em ICCV}, pages 14539--14548, 2021.

\bibitem{patrick2021keeping}
Mandela Patrick, Dylan Campbell, Yuki Asano, Ishan Misra, Florian Metze,
  Christoph Feichtenhofer, Andrea Vedaldi, and Jo{\~a}o~F Henriques.
\newblock Keeping your eye on the ball: Trajectory attention in video
  transformers.
\newblock {\em NeurIPS}, 34, 2021.

\bibitem{perazzi2016benchmark}
Federico Perazzi, Jordi Pont-Tuset, Brian McWilliams, Luc Van~Gool, Markus
  Gross, and Alexander Sorkine-Hornung.
\newblock A benchmark dataset and evaluation methodology for video object
  segmentation.
\newblock In {\em CVPR}, pages 724--732, 2016.

\bibitem{shen2020video}
Wang Shen, Wenbo Bao, Guangtao Zhai, Li~Chen, Xiongkuo Min, and Zhiyong Gao.
\newblock Video frame interpolation and enhancement via pyramid recurrent
  framework.
\newblock {\em IEEE TIP}, 30:277--292, 2020.

\bibitem{shi2021video}
Zhihao Shi, Xiangyu Xu, Xiaohong Liu, Jun Chen, and Ming-Hsuan Yang.
\newblock Video frame interpolation transformer.
\newblock {\em arXiv preprint arXiv:2111.13817}, 2021.

\bibitem{soomro2012ucf101}
Khurram Soomro, Amir~Roshan Zamir, and Mubarak Shah.
\newblock {UCF101}: A dataset of 101 human actions classes from videos in the
  wild.
\newblock {\em arXiv preprint arXiv:1212.0402}, 2012.

\bibitem{sun2018pwc}
Deqing Sun, Xiaodong Yang, Ming-Yu Liu, and Jan Kautz.
\newblock {PWC-Net}: Cnns for optical flow using pyramid, warping, and cost
  volume.
\newblock In {\em CVPR}, pages 8934--8943, 2018.

\bibitem{vaswani2017attention}
Ashish Vaswani, Noam Shazeer, Niki Parmar, Jakob Uszkoreit, Llion Jones,
  Aidan~N Gomez, {\L}ukasz Kaiser, and Illia Polosukhin.
\newblock Attention is all you need.
\newblock In {\em NeurIPS}, pages 5998--6008, 2017.

\bibitem{wang2004image}
Zhou Wang, Alan~C Bovik, Hamid~R Sheikh, and Eero~P Simoncelli.
\newblock Image quality assessment: from error visibility to structural
  similarity.
\newblock {\em IEEE TIP}, 13(4):600--612, 2004.

\bibitem{wu2018video}
Chao-Yuan Wu, Nayan Singhal, and Philipp Krahenbuhl.
\newblock Video compression through image interpolation.
\newblock In {\em ECCV}, pages 416--431, 2018.

\bibitem{xu2019quadratic}
Xiangyu Xu, Li~Siyao, Wenxiu Sun, Qian Yin, and Ming-Hsuan Yang.
\newblock Quadratic video interpolation.
\newblock {\em NeurIPS}, 32, 2019.

\bibitem{xue2019video}
Tianfan Xue, Baian Chen, Jiajun Wu, Donglai Wei, and William~T Freeman.
\newblock Video enhancement with task-oriented flow.
\newblock {\em IJCV}, 127(8):1106--1125, 2019.

\bibitem{yang2020learning}
Fuzhi Yang, Huan Yang, Jianlong Fu, Hongtao Lu, and Baining Guo.
\newblock Learning texture transformer network for image super-resolution.
\newblock In {\em CVPR}, pages 5791--5800, 2020.

\bibitem{zeng2020learning}
Yanhong Zeng, Jianlong Fu, and Hongyang Chao.
\newblock Learning joint spatial-temporal transformations for video inpainting.
\newblock In {\em ECCV}, pages 528--543, 2020.

\bibitem{zhao2019enhanced}
Lei Zhao, Shiqi Wang, Xinfeng Zhang, Shanshe Wang, Siwei Ma, and Wen Gao.
\newblock Enhanced motion-compensated video coding with deep virtual reference
  frame generation.
\newblock {\em IEEE TIP}, 28(10):4832--4844, 2019.

\bibitem{zou2018df}
Yuliang Zou, Zelun Luo, and Jia-Bin Huang.
\newblock {DF-Net}: Unsupervised joint learning of depth and flow using
  cross-task consistency.
\newblock In {\em ECCV}, pages 36--53, 2018.

\end{thebibliography}

\end{document}